\documentclass[letterpaper]{article} % DO NOT CHANGE THIS
\usepackage{aaai25}  % DO NOT CHANGE THIS
\usepackage{times}  % DO NOT CHANGE THIS
\usepackage{helvet}  % DO NOT CHANGE THIS
\usepackage{courier}  % DO NOT CHANGE THIS
\usepackage[hyphens]{url}  % DO NOT CHANGE THIS
\usepackage{graphicx} % DO NOT CHANGE THIS
\usepackage{natbib}  % DO NOT CHANGE THIS AND DO NOT ADD ANY OPTIONS TO IT
\usepackage{caption} % DO NOT CHANGE THIS AND DO NOT ADD ANY OPTIONS TO IT
\usepackage{algorithm}
\usepackage{algorithmic}

\usepackage{multirow}
\usepackage{booktabs}
\usepackage{pifont}
\usepackage[table,xcdraw]{xcolor}

\usepackage{newfloat}
\usepackage{listings}
\DeclareCaptionStyle{ruled}{labelfont=normalfont,labelsep=colon,strut=off} % DO NOT CHANGE THIS
\floatstyle{ruled}
\newfloat{listing}{tb}{lst}{}
\floatname{listing}{Listing}
\title{MUSES: 3D-Controllable Image Generation via Multi-Modal Agent Collaboration}
\author{
    Yanbo Ding\textsuperscript{\rm1, 2}, 
    Shaobin Zhuang\textsuperscript{\rm3, 4}, 
    Kunchang Li\textsuperscript{\rm1, 2, 3}, 
    Zhengrong Yue\textsuperscript{\rm3, 4}, 
    Yu Qiao\textsuperscript{\rm1, 3}, 
    Yali Wang\textsuperscript{\rm1, 3}\textsuperscript{$\dag$}
}
\affiliations{
    \textsuperscript{\rm 1}Shenzhen Key Lab of Computer Vision and Pattern Recognition, Shenzhen Institute of Advanced Technology, Chinese Academy of Sciences\\
    \textsuperscript{\rm 2}School of Artificial Intelligence, University of Chinese Academy of Sciences\\
    \textsuperscript{\rm 3}Shanghai Artificial Intelligence Laboratory\\
    \textsuperscript{\rm 4}Shanghai Jiao Tong University\\
    \{yb.ding, yl.wang\}@siat.ac.cn\\
}

\usepackage{bibentry}
\begin{document}

\twocolumn[{%
\renewcommand\twocolumn[1][]{#1}%
\maketitle

\vspace{-10px}
\begin{center}
    \includegraphics[width=0.95\linewidth]{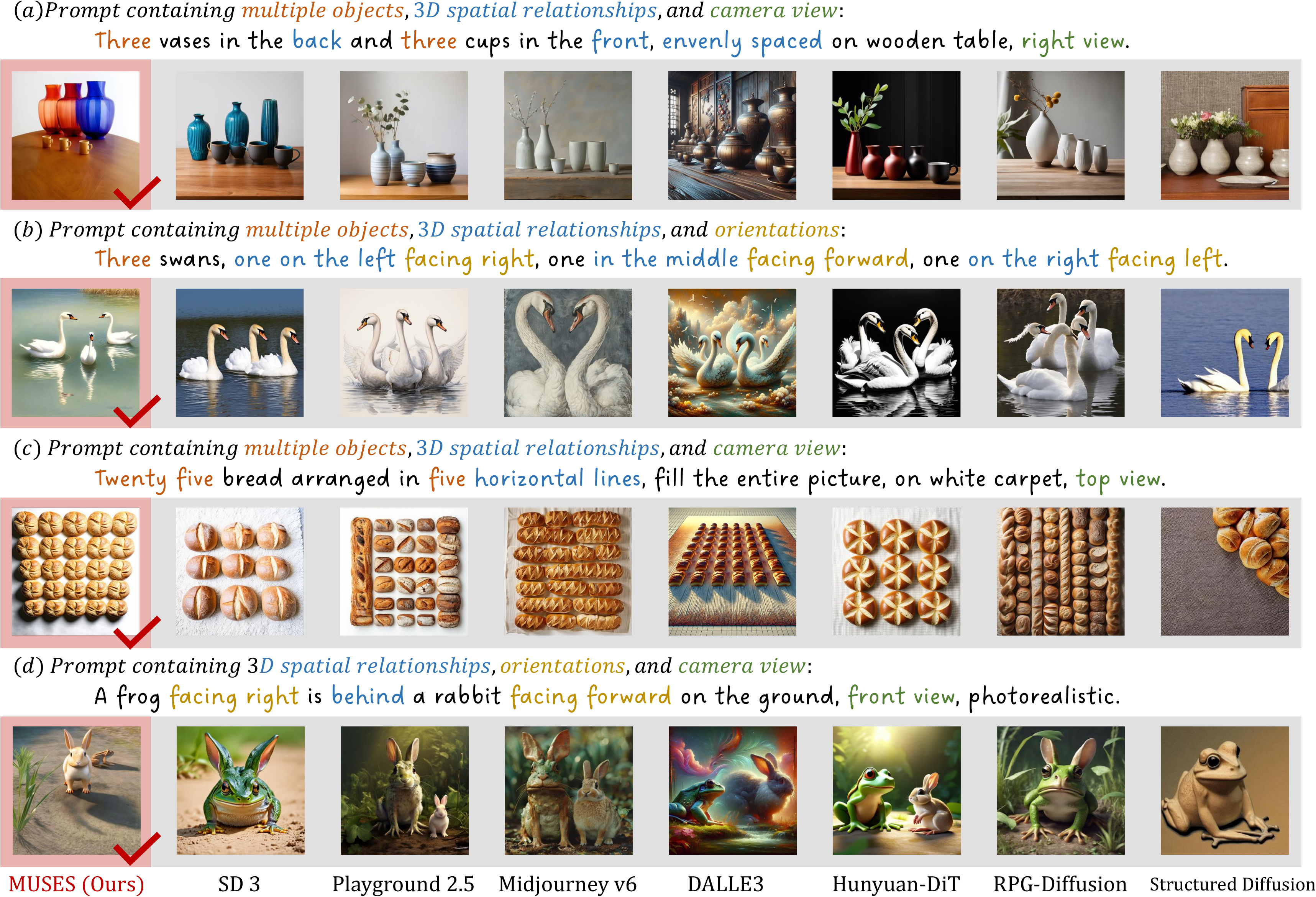}%
    \captionof{figure}{
    \textbf{Comparison Results With Various Methods.} 
    Our MUSES achieves the best, 
    with object numbers highlighted in \textcolor[RGB]{190, 84, 30}{brown}, 
    object orientations in \textcolor[RGB]{200,150,60}{yellow}, 
    3D spatial relationships in \textcolor[RGB]{41, 128, 185}{blue}, 
    and camera views in \textcolor[RGB]{79, 121, 66}{green}, 
    outperforming both open-sourced state-of-the-art methods and commercial API products, such as Stable Diffusion V3,  DALL-E 3, and Midjourney v6.0.}
    \label{fig1: simple_prompts_demos}
    \vspace{6px}
\end{center}%
}]

\begin{figure*}[t]
\centering
\includegraphics[width=1.95\columnwidth]{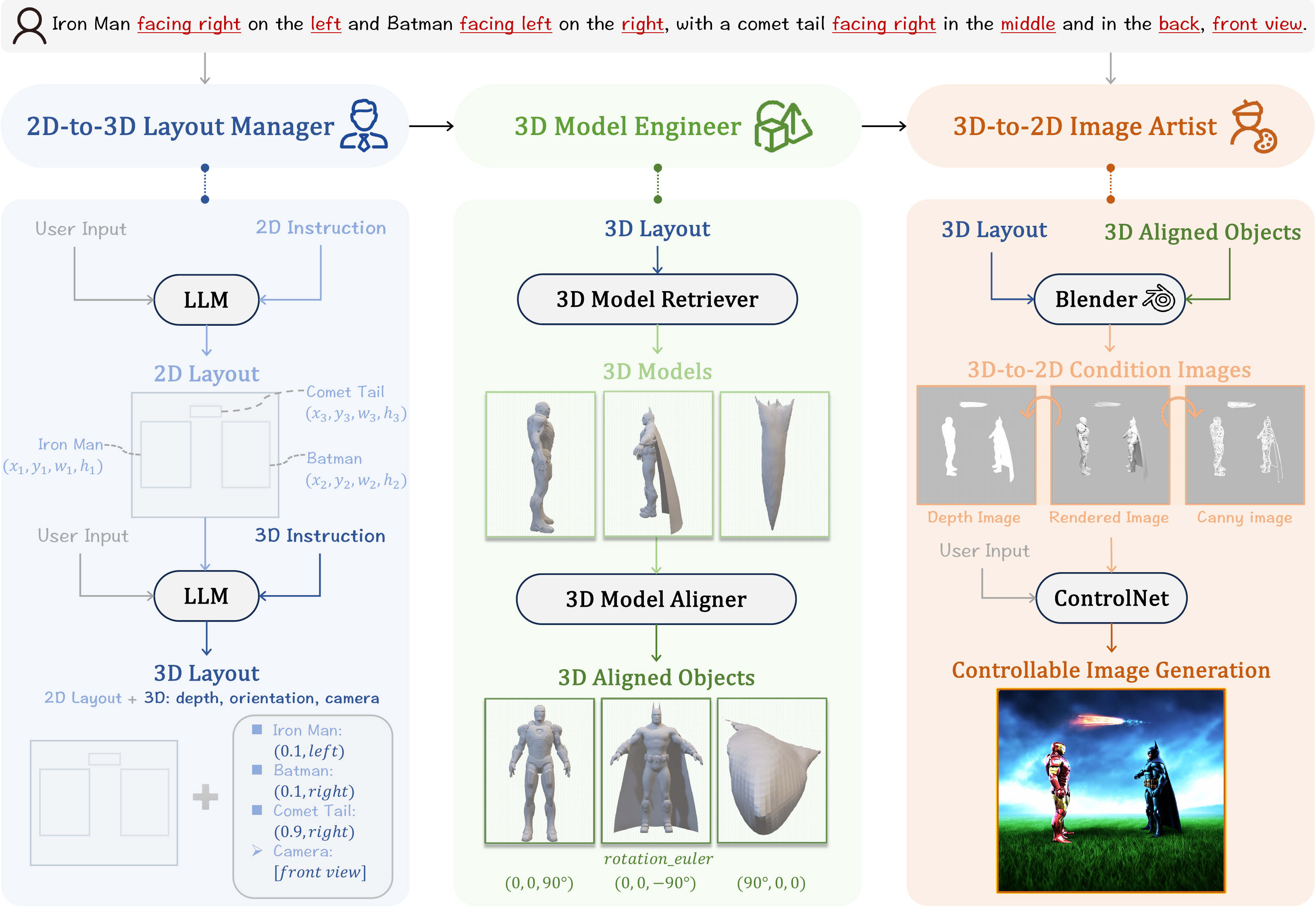}
\caption{\textbf{Overview of our MUSES.} 
Based on the user input, 
Layout Manager first plans a 2D layout and lifts it to a 3D one. 
Then,
Model Engineer acquires 3D models of query objects and aligns them to face the camera. 
Finally, 
Image Artist assembles all the 3D object models into visual conditions that are used for final controllable image generation.}
\vspace{-6px}
\label{fig2: overview}
\end{figure*}

\begingroup
\renewcommand{\thefootnote}{}
\footnote{ \dag \ Corresponding Author.\\
{Copyright \copyright\space 2025
Association for the Advancement of Artificial Intelligence (www.aaai.org).
All rights reserved.}}
\addtocounter{footnote}{-1}
\endgroup

\begin{abstract}
Despite recent advancements in text-to-image generation, 
most existing methods struggle to create images with multiple objects and complex spatial relationships in the 3D world.
To tackle this limitation,
we introduce a generic AI system, 
namely \textbf{MUSES}, 
for 3D-controllable image generation from user queries.
Specifically,
our MUSES develops a progressive workflow with three key components,
including
(1) Layout Manager for 2D-to-3D layout lifting,
(2) Model Engineer for 3D object acquisition and calibration, 
(3) Image Artist for 3D-to-2D image rendering.
By mimicking the collaboration of human professionals, 
this multi-modal agent pipeline facilitates the effective and automatic creation of images with 3D-controllable objects,
through an explainable integration of top-down planning and bottom-up generation. 
Additionally,
existing benchmarks lack detailed descriptions of complex 3D spatial relationships of multiple objects.
To fill this gap, 
we further construct a new benchmark of T2I-3DisBench (3D image scene), 
which describes diverse 3D image scenes with 50 detailed prompts.
Extensive experiments show the state-of-the-art performance of MUSES on both T2I-CompBench and T2I-3DisBench,
outperforming recent strong competitors such as DALL-E 3 and Stable Diffusion 3.
These results demonstrate a significant step forward for MUSES in bridging natural language, 2D image generation, and 3D world.
\end{abstract}

\begin{links}
    \link{Code}{https://github.com/DINGYANB/MUSES}
    \link{Dataset}{https://huggingface.co/yanboding/MUSES}
\end{links}

\section{Introduction}

Text-to-image generation \citep{sd,dalle,midjourney} is rapidly evolving in quality.
However,
such generation often struggles with detailed user queries of multiple objects in complex scenes.
Several approaches have emerged to address this by compositional text-to-image synthesis 
\citep{rpg-diffusion,exct,layoutgpt,struct-diffusion}.
Unfortunately,
they fail to accurately control 3D contents like object orientation and camera view,
even though
our real world is inherently three-dimensional.
As shown in Fig. \ref{fig1: simple_prompts_demos},
when the prompt is ``a frog facing right is behind a rabbit facing forward", 
existing methods collapse with unsatisfactory 3D arrangements. 
This raises a fundamental question:
can we create images with precise 3D control to better simulate our world?

To answer this question,
we draw inspiration from the workflow of 3D professionals.
We observe that
creating such images typically involves three key stages: 
scene layout planning,
3D objects design,
and image rendering \citep{physically}.
This highlights 
the need for developing a systematic framework of 3D-controllable image creation, 
rather than relying on a single generation model.
Therefore,
we propose a generic AI system that automatically and precisely creates images with 3D controllable objects.
We name it \textbf{MUSES},
since we consider human designers as our ``Muses",
and mimic their workflows through a collaborative pipeline of multi-modal agents.

Our MUSES system, as depicted in Fig. \ref{fig2: overview}, comprises three key agents that progressively achieve 3D-controllable image generation:
(1) \textit{Layout Manager}.
First,
we employ a Large Language Model (e.g., Llama3 \citep{llama3}) to plan and assign 3D object locations based on user queries.
Our innovative 2D-to-3D layout lifting paradigm first generates a 2D layout by in-context learning, 
then elevates it to a 3D layout via chain-of-thought reasoning.
(2) \textit{Model Engineer}.
After obtaining the 3D layout,
we introduce a model engineer to acquire 3D models of queried objects.
To enhance its robustness,
we design a flexible retriever that gathers 3D models
through a decision tree approach, combining self-collected model shop retrieval, online search, and text-to-3D generation.
Furthermore,
to ensure orientation alignment with the planned 3D layout,
we develop a novel aligner to calibrate object orientation
by face-camera-view identification with CLIP \citep{clip}.
(3) \textit{Image Artist}.
Finally,
we introduce an image artist to render 3D-controllable images.
The 3D-aligned objects and their layouts are fed into Blender \citep{blender},
which accurately assembles all the objects into 3D-to-2D condition images.
These conditions are then 
used with ControlNet \citep{controlnet} to generate the final image.

Our contributions are threefold.
\textit{First},
our MUSES is the first AI system for 3D-controllable image generation,
to our best knowledge. 
It enables precise control over object properties 
such as object count, orientation, 3D spatial relationships, 
and camera view, 
potentially bridging the gap between image generation and world simulation.
\textit{Second},
MUSES is a distinct multi-agent collaboration for 3D-controllable image generation. 
Each agent of MUSES is an insightful and novel integration of multi-modal agents,
allowing for top-down planning and bottom-up generation with robust control of 3D information.
\textit{Finally},
since
existing benchmarks lack detailed descriptions of complex 3D information like object orientation and camera view, 
we further construct a new benchmark of T2I-3DisBench (3D image scene),
which consists of 50 prompts involving multiple objects with
diverse object orientations, 
3D spatial relationships, 
and camera views across various complex 3D scenes.
%six types of 3D spatial relationships, 
%six types of object orientations, 
%and four types of camera views.
Extensive experiments demonstrate the superiority of MUSES on both T2I-CompBench and our T2I-3DisBench,
where
it consistently outperforms state-of-the-art competitors of both open-source models and commercial API products,
including Stable Diffusion v3 \citep{sd3}, DALL-E 3 \citep{dalle3} and Midjourney v6.0 \citep{midjourney},
in terms of precise 3D control in image generation.

\begin{figure*}[t]
\centering
\includegraphics[width=2.0\columnwidth]{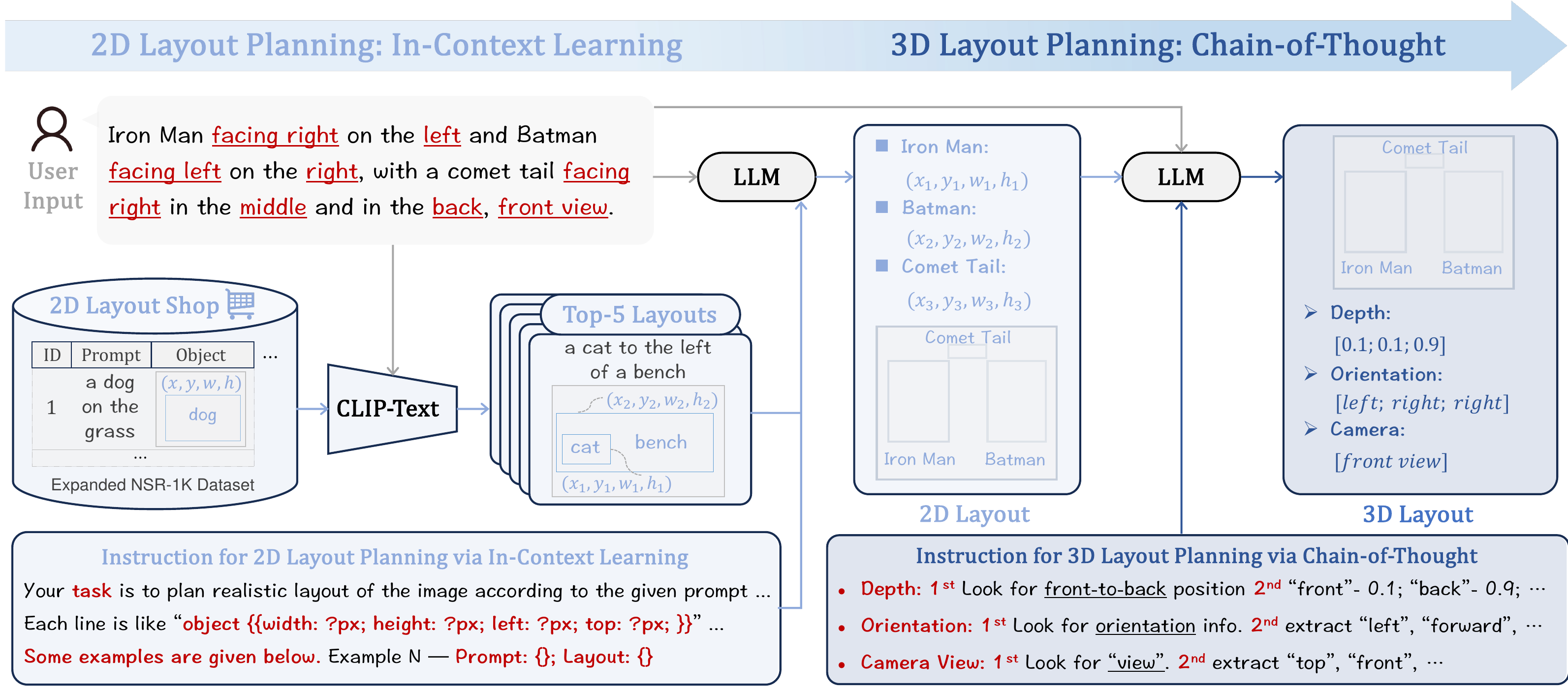}
\caption{\textbf{2D-to-3D Layout Manager.} 
First, based on the user input, our layout manager employs the LLM to plan 2D layout through In-context Learning. 
Then, it lifts the 2D layout to 3D space via Chain of Thought reasoning.}
\label{fig3: layout_manager}
\vspace{-6px}
\end{figure*}

\section{Related Work}
\noindent \textbf{Controllable Image Generation.}
Before diffusion \citep{ddpm, diffusion}, 
GAN-based \citep{gan} methods 
such as ControlGAN \citep{controlgan} and AttnGAN \citep{attengan}, 
incorporate text features via attention \citep{attention} modules to guide image generation.
In recent years, 
Stable Diffusion series \citep{sdxl, sd2, sd} have dominated the text-to-image generation market.
Given that text-based control is insufficient for precise image generation, 
ControlNet \citep{controlnet}
introduced additional fine-grained conditions (e.g., depth maps). 
Additionally, 
models such as Structured-Diffusion \citep{struct-diffusion} and Attn-Exct \citep{exct} fuse linguistic structures or attention-based semantic guidance into the diffusion process.
Approaches like LayoutGPT \citep{layoutgpt} and LMD \citep{lmd} use LLM to plan 2D layouts (bounding boxes), 
while RPG \citep{rpg-diffusion} plans and assigns regions based on the user input.
However, 
existing methods struggle to control 3D properties of objects.
We instead plan 3D layouts and incorporate 3D models and simulations to achieve 3D controllable image generation.

\noindent \textbf{LLM-Based Agents.}
LLMs like GPT series \citep{brown2020language, gpt4} and Llama series \citep{llama2, touvron2023llama} have revolutionized natural language processing \citep{chowdhary2020natural}.
MLLMs like LLaVA \citep{llava} and InternVL \citep{internvl} have enabled impressive performance on visual tasks. 
The combination of LLMs and MLLMs in multi-agent systems achieves remarkable success across various domains, 
including 
visual understanding \citep{visiongpt, visualgpt}, 
gaming \citep{tom,mindagent}, 
software development \citep{autogen, chatdev}, 
video generation \citep{mora, worldgpt}, 
and autonomous driving \citep{chatsim, multi}.
We focus on image generation via multi-agent collaboration.
DiffusionGPT \citep{diffusiongpt} uses LLM to select models in image generation.
SLD \citep{sld} uses LLM for self-correcting the generated image. 
CompAgent \citep{compAgent} uses LLM to coordinate the image generation process into sub-steps.
Unlike these works, 
we use LLM to plan and lift 2D layouts to 3D.

\section{Method}

In this section, 
we introduce our MUSES for 3D controllable image generation.
As shown in Fig. \ref{fig2: overview}, 
it is a generic AI system with a distinct multi-modal agent collaboration pipeline.
Specifically,
our MUSES consists of three collaborative agents including
Layout Manager for 2D-to-3D layout lifting,
Model Engineer for 3D object acquisition and calibration,  
Image Artist for 3D-to-2D image rendering. 

\subsection{Lifting: 2D-to-3D Layout Manager}

To achieve precise 3D control,
we first plan a 3D layout according to the user input.
Specifically,
we employ LLM (e.g., Llama3 \citep{llama3}) as a layout manager due to its great power of linguistic understanding.
To alleviate planning difficulty,
we design a 2D-to-3D lifting paradigm for the progressive transition from 2D to 3D layout in Fig. \ref{fig3: layout_manager}.

\textbf{2D Layout Planning via In-Context-Learning.} 
We start by planning the 2D position of the objects.
Apparently,
asking LLM directly to generate an object layout is not ideal, 
as it may struggle with managing bounding boxes in images.
Hence,
we leverage In-Context Learning \citep{icl},
allowing LLM to follow instructions with a few examples from a 2D layout shop.
We use the NSR-1K dataset \citep{layoutgpt} as the 2D layout shop,
since it
contains over 40,000 entries with diverse prompts, 
objects, and the corresponding bounding boxes. 
As the original NSR-1K dataset lacks layouts in 3D scenes and complex scenes,
we manually designed and add some 3D layouts (details in Appendix).
First,
we use CLIP \citep{clip} text encoder to compare similarities of user input and text prompts in the shop.
Then,
we select the top five 2D layouts with the highest similarity scores. 
Finally,
we feed these contextual layouts along with instructions to LLM.
Such a comprehensive approach results in a contextually relevant 2D layout, 
forming the foundation for the subsequent 3D lifting.

\noindent\textbf{3D Layout Planning via Chain-of-Thought.}
Unlike previous approaches \citep{layoutgpt, layoutllm, lmd} 
that directly use 2D layout for image generation, 
we lift our 2D layout to 3D space.
Specifically,
With 2D layout and user input,
we further ask LLM to plan 3D attributes, 
including 
depth,
orientation, 
and camera view. 
To enhance complex planning capability,
we design a chain-of-thought (CoT) prompt \citep{cot} 
for step-by-step reasoning on each attribute.
Taking depth as an example,
the first step instruction is to look for front-to-back position relationships in user input.
Based on such relationships,
the second-step instruction is to assign a depth value to each object,
(e.g.,
``A is in front of B": A's depth is set to 0.1 and B's depth is set to 0.9).
The instructions for orientation and camera view are similar.
We list all instructions in Appendix. 
Via such a concise manager,
we can accurately exploit the 3D layout in user input for subsequent 3D simulation in Blender.

\begin{figure*}[t]
\centering
\includegraphics[width=1.95\columnwidth]{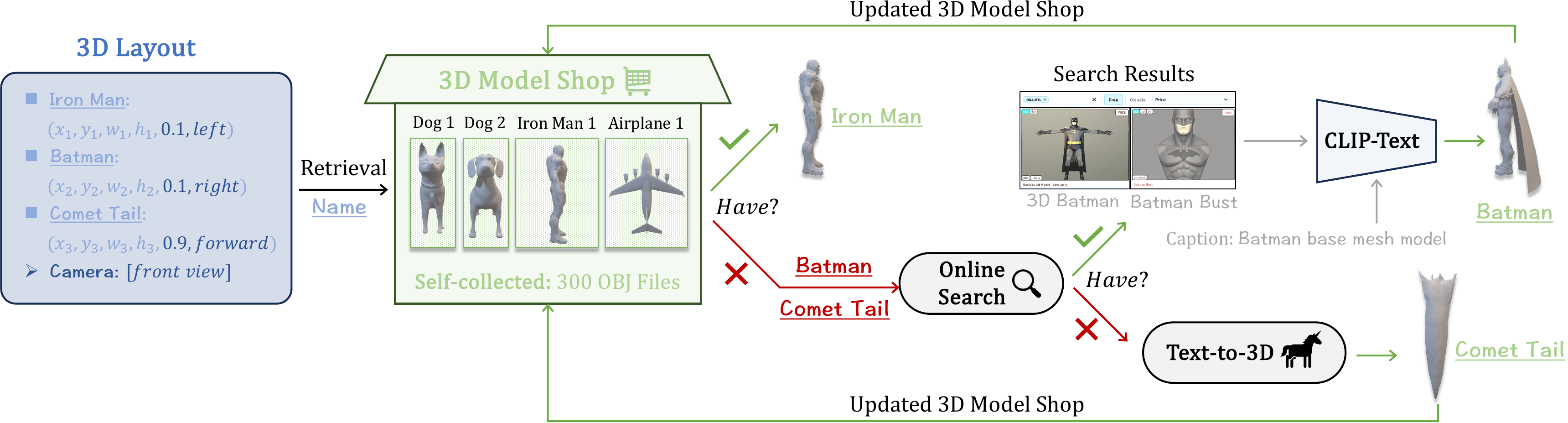}
\caption{\textbf{3D Model Retriever.} 
We develop a retrieve-generate decision tree that automatically acquires 3D objects specified in the 3D layout from a self-collected model shop,
based on a concise decision process of online search and text-to-3D generation.}
\label{fig4: retriever}
\vspace{-6px}
\end{figure*}

\subsection{Calibrating: 3D Model Engineer}

After planning the 3D layout, the next step is to acquire specified 3D models.
Specifically,
we introduce a Model Engineer which comprises two key roles:
Model Retriever for acquiring 3D models in the 3D layout, 
Model Aligner for calibrating the orientations of 3D models to face camera.

\textbf{3D Model Retriever: Retrieve-Generate Decision Tree}. 
To obtain 3D models with efficiency and robustness,
we design a concise Model Retriever via the decision tree of retrieval and generation in Fig. \ref{fig4: retriever}.
Our motivation is that, 
3D models from the internet have higher quality
compared to those generated by text-to-3D.
Hence,
we prioritize the retrieval of existing 3D models in the decision tree, 
which not only enhances the overall visual quality of 3D models, but also reduces computational load in text-to-3D synthesis. 

Specifically,
our model retriever is based on a 3D model shop (300 3D models of 230 object categories) that is self-built in an online fashion.
\textit{First},
the model retriever looks for 3D models of queried objects from the current shop,
using the object name as the search key.
\textit{Second},
if it cannot find any 3D models of a queried object from this shop (e.g., Batman in Fig. \ref{fig4: retriever}),
it will search online,
e.g., 
on the professional website (https://www.cgtrader.com).
For each object,
there may exist numerous 3D models from such a website.
Hence,
we perform CLIP text encoder to calculate the similarity between the object name and the online item title,
and select the 3D model with the highest similarity.
\textit{Third},
if online search also fails in finding suitable 3D models (e.g., Comet Halley in Fig. \ref{fig4: retriever}),
it will employ a text-to-3D generation model like Shap-E \citep{shape} or 3DTopia \citep{3dtopia} to synthesize corresponding 3D model. 
\textit{Finally},
we add the newly found object models to our 3D model shop to enhance shop diversity and versatility of future usage.
In such a manner,
we obtain 3D models of query objects and remain up-to-date with the latest available high-quality models,
ensuring both efficiency and robustness.

\begin{figure*}[t]
\centering
\includegraphics[width=1.95\columnwidth]{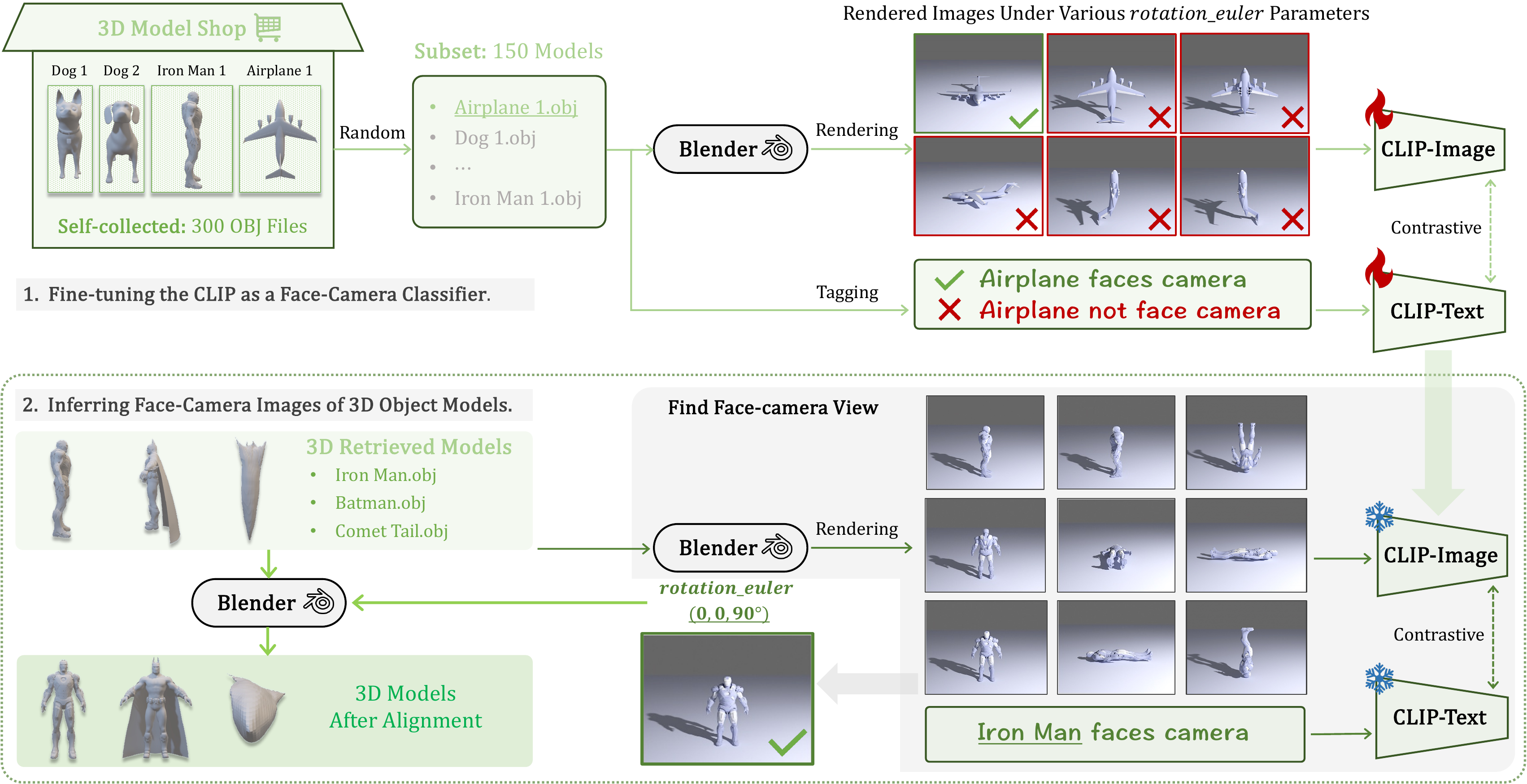}
\caption{\textbf{3D Model Aligner.} 
It aligns 3D models with face-camera orientation,
ensuring that the final orientation conforms to the planned 3D layout.
First,
we fine-tune CLIP as a Face-Camera Classifier, by a training set generated from our 3D model shop.
Then,
we use fine-tuned CLIP to identify the face-camera image of each object, aligning its 3D model to face the camera.}
\label{fig5: aligner}
\vspace{-6px}
\end{figure*}

\textbf{3D Model Aligner: Face-Camera Calibration}.
As 3D models are acquired from internet or generation,
their orientations may not align with the expected ones in the planned 3D layout. 
To address this,
we introduce a novel 3D Model Aligner,
which can calibrate orientations of 3D models to face camera,
for further usage along with 3D layouts.
We propose to fine-tune CLIP as a binary classifier (Fig. \ref{fig5: aligner}) for its strong generalization capacity by large-scale pretraining. 

\textit{Fine-tuning CLIP as a Face-Camera Classifier.} 
First,
we need to prepare the fine-tuning dataset.
We randomly select 150 3D models from our 3D shop and import them into Blender with a standardized environment.
For each 3D model,
we perform multi-view rendering to generate a set of 2D images from various views,
with different rotation\_euler parameters in Blender.
Then,
we annotate images by tagging the description of ``object name (faces / not face) camera''. 
To increase data diversity across 3D geometries,
we randomly select 5 no-face-camera images as negative samples for each 3D model.
To balance positive and negative samples during fine-tuning,
we make the face-camera image as 5 copies for each 3D model.
This results in a training set with 1500 image-text pairs,
which are used for fine-tuning CLIP as a Face-Camera Classifier by contrastive language-image learning. 
After fine-tuning, 
we test it with an extra 1500 images (50\% face camera, 50\% not) from other 150 3D models. 
All test images are correctly classified.

\textit{Inferring Face-Camera Image of 3D Object Models.} 
%We feed 1500 image-text pairs generated by the above process to fine-tune the CLIP.
%The model learns to maximize similarities between image embedding of objects facing the camera and the text embedding of "a/an (object name) faces camera",
%while 
%minimizing similarities with "a/an (object name) not face camera". 
During the inference, 
we import 3D models of queried objects (from model retriever) into the same Blender environment used in training dataset generation.
For each 3D model,
we perform multi-view rendering to generate a comprehensive set of 2D images from various views.
Then,
we use the fine-tuned CLIP to identify the face-camera image of each 3D model,
by comparing similarities between the rendered images and the text ``object name faces camera".
Based on this image,
we can effectively align 3D object models to face the camera through the configuration of the rotation parameter in Blender,
which is used to generate correctly orientated objects in the subsequent 3D-to-2D image generation.

\subsection{Rendering: 3D-to-2D Image Artist}
So far,
we have obtained a 3D layout and aligned 3D models in the user query.
Given these 3D materials,
we next introduce a concise image artist to create the 3D-controllable image,
based on a 3D-to-2D rendering as shown in Fig. \ref{fig2: overview}.
More Blender rendering outputs can be found in Appendix.
%%%%%%%%%%%%%%%%%
%\textbf{3D-to-2D Condition Image Rendering.}
First,
we assemble all the 3D object models into a complete scene based on the 3D layout.
To ensure consistent and accurate 3D scene composition,
we develop a comprehensive setting of parameter configuration in Blender such as settings of rendering environment, camera, and object (detailed Blender configurations are in Appendix).
Once the 3D scene is fully assembled, 
we use the engine, ``CYCLES", to convert the 3D scene into a 2D image. 
To enhance the control over the final image generation, 
we process this 2D rendering into two condition images,
including
(1) Depth Map by Blender's Z-pass rendering \citep{simple}, representing 3D spatial relationships.
(2) Canny Edge by OpenCV \citep{opencv}, highlighting contours.
Finally,
we leverage these 3D-to-2D condition images as fine-grained control,
and use advanced image generation techniques like ControlNet \citep{controlnet},
to generate the final image with the user input.
Via such a concise image artist,
our MUSES can flexibly generate a 3D-controllable image that accurately reflects both 3D spatial details and semantic contents of the user input.

\begin{table*}[ht]
    \centering
    \setlength{\tabcolsep}{1.8mm}
    \fontsize{9pt}{11pt}\selectfont
    \begin{tabular}{lccccccc}
        \toprule
        \multirow{2}{*}{\textbf{Method}} 
        & \multicolumn{3}{c}{\textbf{Attribute Binding}} 
        & \multicolumn{3}{c}{\textbf{Object Relationship}} 
        & \multirow{2}{*}{\textbf{Numeracy$\uparrow$}} \\
        
        \cmidrule(r){2-4} \cmidrule(r){5-7}
         & \textbf{Color$\uparrow$} 
         & \textbf{Shape$\uparrow$} 
         & \textbf{Texture$\uparrow$} 
         & \textbf{2D-Spatial$\uparrow$} 
         & \textbf{3D-Spatial$\uparrow$} 
         & \textbf{Non-Spatial$\uparrow$} \\
        \midrule
        
        LayoutGPT \citep{layoutgpt} & 0.2921 & 0.3716 & 0.3310 & 0.1153 & 0.2607 & 0.2989 & 0.4193 \\
        Structured Diffusion \citep{struct-diffusion} & 0.4990 & 0.4218 & 0.4900 & 0.1386 & 0.2952 & 0.3111 & 0.4562 \\
        Attn-Exct \citep{exct}
        & 0.6400 & 0.4517 & 0.5963 & 0.1455 & - & 0.3109 & - \\
        GORS \citep{t2i-bench}
        & 0.6603 & 0.4785 & 0.6287 & 0.1815 & - & 0.3193 & - \\
        RPG-Diffusion \citep{rpg-diffusion}
        % 0.6024, 0.4597, 0.5326, 0.3493, (0.4269)
        & 0.6024 & 0.4597 & 0.5326 & 0.2115 & 0.3587 & 0.3104 & 0.4968 \\
        CompAgent \citep{compAgent}
        & 0.7400 & 0.6305 & 0.7102 & 0.3698 & - & 0.3104 & - \\
        
        \midrule
        % SD v1.4 \citep{sd} & 0.3765 & 0.3576 & 0.4156 & 0.1350 & 0.3336 & 0.3079 & 0.4418 \\
        % SD v2 \citep{sd2} & 0.5065 & 0.4221 & 0.4922 & 0.1703 & 0.3205 & 0.3096 & 0.4849 \\
        SDXL \citep{sdxl} & 0.6369 & 0.5408 & 0.5637 & 0.2032 & 0.3438 & 0.3110 & 0.5145 \\
        PixArt-$\alpha$ \citep{pixart} & 0.6886 & 0.5582 & 0.7044 & 0.2082 & 0.3530 & 0.3179 & 0.5001 \\
        Playground v2.5 \citep{playgroundv25} & 0.6381 & 0.4790 & 0.6297 & 0.2062 & 0.3816 & 0.3108 & 0.5329 \\
        Hunyuan-DiT \citep{hunyuandit} & 0.6342 & 0.4641 & 0.5328 & 0.2337 & 0.3731 & 0.3063 & 0.5153 \\
        DALL-E 3 \citep{dalle3} & 0.7785 & 0.6205 & 0.7036 & 0.2865 & - & 0.3003 & - \\
        SD v3 \citep{sd3} & 0.8085 & 0.5793 & 0.7317 & 0.3144 & 0.4026 & 0.3131 & 0.6088 \\
        \midrule
        % \textbf{MUSES(Ours)} & 0.7668 & 0.5652 & 0.7106 & 0.3544 & 0.3893 & 0.3214 & 0.5620 \\
        % {\textbf{MUSES(Ours)}} 
        % & {\textbf{0.7668}} 
        % & {\textbf{0.5762}} 
        % & {\textbf{0.7206}} 
        % & {\textbf{0.3544}}
        % & {\textbf{0.3893}}
        % & {\textbf{0.3214}} 
        % & {\textbf{0.5673}} \\
        % \rowcolor{gray!30}
        {\textbf{MUSES (Ours)}} 
        & {\textbf{0.8726}} 
        & {\textbf{0.6812}} 
        & {\textbf{0.8081}} 
        & {\textbf{0.4756}}
        & {\textbf{0.4639}}
        & {\textbf{0.3226}} 
        & {\textbf{0.7720}} \\
        \bottomrule
    \end{tabular}
    \caption{\textbf{Evaluation Results on T2I-CompBench.} Our MUSES demonstrates the best performance on attribute binding, object relationship, and object count, outperforming SOTA methods, including multi-agent specialized methods, and generic models.}
    \label{table: ti2-combench-evaluation}
\end{table*}

\begin{table*}[ht]
    \centering
    \setlength{\tabcolsep}{0.9mm}
    \fontsize{9pt}{11pt}\selectfont
    \begin{tabular}{lcccccc}
        \toprule
        \textbf{Method} &  \textbf{Average Score} $\uparrow$ & \textbf{Object Count} $\uparrow$ & \textbf{Orientation} $\uparrow$ & \textbf{3D Spatial Relationship} $\uparrow$ & \textbf{Camera View} $\uparrow$ \\
        \midrule

        Structured Diffusion \citep{struct-diffusion} & 0.1862 (0.15) & 0.2160 (0.14) & 0.1647 (0.08) & 0.1773 (0.21) & 0.1866 (0.18) \\
        % \rowcolor{gray!30}
        % Structured Diffusion \citep{struct-diffusion} & 0.3274 (0.15) & 0.3395 (0.14) & 0.2933 (0.08) & 0.4130 (0.21) & 0.2637 (0.18) \\
        
        RPG-Diffusion \citep{rpg-diffusion} & 0.2753 (0.18) & 0.3209 (0.18) & 0.2533 (0.15) & 0.3195 (0.25) & 0.2075 (0.12) \\
        % \rowcolor{gray!30}
        % RPG-Diffusion \citep{rpg-diffusion} & 0.3167 (0.18) & 0.3475 (0.18) & 0.3041 (0.15) & 0.3410 (0.25) & 0.2742 (0.12) \\
        
        LayoutGPT \citep{layoutgpt} & 0.2348 (0.14) & 0.2937 (0.14) & 0.2058 (0.12) & 0.2783 (0.18) & 0.1613 (0.10) \\
        % \rowcolor{gray!30}
        % LayoutGPT \citep{layoutgpt} & 0.4370 (0.14) & 0.4204 (0.14) & 0.3916 (0.12) & 0.5258 (0.18) & 0.4100 (0.10) \\
        
        \midrule

        Playground v2.5 \citep{playgroundv25}  & 0.2344 (0.19) & 0.3587 (0.19) & 0.1887 (0.17) & 0.2071 (0.28) & 0.1830 (0.15) \\
        % \rowcolor{gray!30}
        % Playground v2.5 \citep{playgroundv25}  & 0.2381 (0.19) & 0.3590 (0.19) & 0.1550 (0.17) & 0.2639 (0.28) & 0.1746 (0.15) \\
    
        DALL-E 3 \citep{dalle3}  & 0.2627 (0.23) & 0.3013 (0.15) & 0.2363 (0.18) & 0.2370 (0.29) & 0.2757 (0.29) \\
        % \rowcolor{gray!30}
        % DALL-E 3 \citep{dalle3}  & 0.2606 (0.23) & 0.2930 (0.15) & 0.2010 (0.18) & 0.2897 (0.29) & 0.2587 (0.29) \\
        
        Hunyuan-DiT \citep{hunyuandit}  & 0.2780 (0.22) & 0.3496 (0.19) & 0.2517 (0.21) & 0.2598 (0.29) & 0.2510 (0.18) \\
        % \rowcolor{gray!30}
        % Hunyuan-DiT \citep{hunyuandit}  & 0.2745 (0.22) & 0.3773 (0.19) & 0.1657 (0.21) & 0.3427 (0.29) & 0.2123 (0.18) \\
        
        Midjourney v6.0 \citep{midjourney}  & 0.3760 (0.35) & 0.4470 (0.39) & 0.3438 (0.27) & 0.3518 (0.36) & 0.3613 (0.38) \\
        % \rowcolor{gray!30}
        % Midjourney v6.0 \citep{midjourney}  & 0.4314 (0.35) & 0.5240 (0.39) & 0.3650 (0.27) & 0.5207 (0.36) & 0.3160 (0.38) \\

        SD v3 \citep{sd3}  & 0.4206 (0.36) & 0.5383 (0.42) & 0.3303 (0.26) & 0.4600 (0.40) & 0.3537 (0.33) \\

        % \rowcolor{gray!30}
        % Stable Diffusion V3 \citep{sd3}  & 0.4950 (0.36) & 0.6473 (0.42) & 0.3630 (0.26) & 0.5679 (0.40) & 0.4017 (0.33) \\

        \midrule
        % \textbf{MUSES(Ours)}  & \textbf{0.5192 (0.62)} & \textbf{0.6180 (0.61)} & \textbf{0.4610 (0.68)} & \textbf{0.5647 (0.62)} & \textbf{0.4330 (0.57)} \\

        % \rowcolor{gray!30}
        \textbf{MUSES (Ours)}  & \textbf{0.6156 (0.62)} & \textbf{0.7488 (0.61)} & \textbf{0.4709 (0.68)} & \textbf{0.7207 (0.62)} & \textbf{0.5220 (0.57)} \\

        \bottomrule
    \end{tabular}
    \caption{\textbf{Evaluation Results on our T2I-3DisBench.} Our MUSES consistently outperforms other methods across all metrics. Each cell displays both the InternVL-VQA metric value and the corresponding score from user evaluation (in parentheses).}
    \label{table: complex-prompts-evaluation}
\end{table*}

\begin{table*}[ht!]
    \centering
    \setlength{\tabcolsep}{0.9mm}
    \fontsize{9pt}{11pt}\selectfont
    \begin{tabular}{lcccccccccccc}
        \toprule
        \textbf{Component} & \multicolumn{12}{c}{\textbf{Choice}} \\
        \midrule
        Object Depth Planning & {\ding{52}} & {\ding{56}} & \textcolor[RGB]{192, 192, 192}{\ding{52}} & \textcolor[RGB]{192, 192, 192}{\ding{52}} & \textcolor[RGB]{192, 192, 192}{\ding{52}} & \textcolor[RGB]{192, 192, 192}{\ding{52}} & \textcolor[RGB]{192, 192, 192}{\ding{52}} & \textcolor[RGB]{192, 192, 192}{\ding{52}} & \textcolor[RGB]{192, 192, 192}{\ding{52}} & {\ding{56}} & \textcolor[RGB]{192, 192, 192}{\ding{52}} & {\ding{56}} \\

        Object Orientation Planning & {\ding{52}} & \textcolor[RGB]{192, 192, 192}{\ding{52}} & {\ding{56}} & \textcolor[RGB]{192, 192, 192}{\ding{52}} & \textcolor[RGB]{192, 192, 192}{\ding{52}} & \textcolor[RGB]{192, 192, 192}{\ding{52}} & \textcolor[RGB]{192, 192, 192}{\ding{52}} & \textcolor[RGB]{192, 192, 192}{\ding{52}} & \textcolor[RGB]{192, 192, 192}{\ding{52}} & {\ding{56}} & \textcolor[RGB]{192, 192, 192}{\ding{52}} & {\ding{56}} \\

        Camera View Planning & {\ding{52}} & \textcolor[RGB]{192, 192, 192}{\ding{52}} & \textcolor[RGB]{192, 192, 192}{\ding{52}} & {\ding{56}} & \textcolor[RGB]{192, 192, 192}{\ding{52}} & \textcolor[RGB]{192, 192, 192}{\ding{52}} & \textcolor[RGB]{192, 192, 192}{\ding{52}} & \textcolor[RGB]{192, 192, 192}{\ding{52}} & \textcolor[RGB]{192, 192, 192}{\ding{52}} & {\ding{56}} & \textcolor[RGB]{192, 192, 192}{\ding{52}} & {\ding{56}} \\

        Retrieve-Generate Decision Tree & {\ding{52}} & \textcolor[RGB]{192, 192, 192}{\ding{52}} & \textcolor[RGB]{192, 192, 192}{\ding{52}} & \textcolor[RGB]{192, 192, 192}{\ding{52}} & {\ding{56}} & \textcolor[RGB]{192, 192, 192}{\ding{52}} & \textcolor[RGB]{192, 192, 192}{\ding{52}} & \textcolor[RGB]{192, 192, 192}{\ding{52}} & \textcolor[RGB]{192, 192, 192}{\ding{52}} & \textcolor[RGB]{192, 192, 192}{\ding{52}} & {\ding{56}} & {\ding{56}} \\

        Face-Camera Calibration & {\ding{52}} & \textcolor[RGB]{192, 192, 192}{\ding{52}} & \textcolor[RGB]{192, 192, 192}{\ding{52}} & \textcolor[RGB]{192, 192, 192}{\ding{52}} & \textcolor[RGB]{192, 192, 192}{\ding{52}} & {\ding{56}} & \textcolor[RGB]{192, 192, 192}{\ding{52}} & \textcolor[RGB]{192, 192, 192}{\ding{52}} & \textcolor[RGB]{192, 192, 192}{\ding{52}} & \textcolor[RGB]{192, 192, 192}{\ding{52}} & {\ding{56}} & {\ding{56}} \\

        CLIP Fine-tuning During Calibration & {\ding{52}} & \textcolor[RGB]{192, 192, 192}{\ding{52}} & \textcolor[RGB]{192, 192, 192}{\ding{52}} & \textcolor[RGB]{192, 192, 192}{\ding{52}} & \textcolor[RGB]{192, 192, 192}{\ding{52}} & {\ding{56}} & {\ding{56}} & \textcolor[RGB]{192, 192, 192}{\ding{52}} & \textcolor[RGB]{192, 192, 192}{\ding{52}} & \textcolor[RGB]{192, 192, 192}{\ding{52}} & {\ding{56}} & {\ding{56}} \\

        Multiple Control Scales & {\ding{52}} & \textcolor[RGB]{192, 192, 192}{\ding{52}} & \textcolor[RGB]{192, 192, 192}{\ding{52}} & \textcolor[RGB]{192, 192, 192}{\ding{52}} & \textcolor[RGB]{192, 192, 192}{\ding{52}} & \textcolor[RGB]{192, 192, 192}{\ding{52}} & \textcolor[RGB]{192, 192, 192}{\ding{52}} & {\ding{56}} & \textcolor[RGB]{192, 192, 192}{\ding{52}} & \textcolor[RGB]{192, 192, 192}{\ding{52}} & \textcolor[RGB]{192, 192, 192}{\ding{52}} & {\ding{56}} \\
        
        3D Layout Shop Expansion & {\ding{52}} & \textcolor[RGB]{192, 192, 192}{\ding{52}} & \textcolor[RGB]{192, 192, 192}{\ding{52}} & \textcolor[RGB]{192, 192, 192}{\ding{52}} & \textcolor[RGB]{192, 192, 192}{\ding{52}} & \textcolor[RGB]{192, 192, 192}{\ding{52}} & \textcolor[RGB]{192, 192, 192}{\ding{52}} & \textcolor[RGB]{192, 192, 192}{\ding{52}} & {\ding{56}} & \textcolor[RGB]{192, 192, 192}{\ding{52}} & \textcolor[RGB]{192, 192, 192}{\ding{52}} & {\ding{56}} \\
        \midrule

        % Results on T2I-CompBench & \textbf{0.3893} & 0.3647 & 0.3816 & 0.3893 & 0.3536 & 0.3567 & 0.3640 & 0.3517 & 0.3439 & 0.3471 & 0.3406 \\

        Results on T2I-CompBench & \textbf{0.4639} & 0.4127 & 0.4609 & 0.4636 & 0.4234 & 0.3828 & 0.4289 & 0.3893 & 0.3925 & 0.3641 & 0.3427  & 0.3093 \\

        % Results on T2I-3DisBench & \textbf{0.5192} & 0.3852 & 0.4025 & 0.4580 & 0.3837 & 0.4988 & 0.4437 & 0.3455 & 0.3558 & 0.3301 & 0.2980 \\

        Results on T2I-3DisBench & \textbf{0.6156} & 0.4981 & 0.4938 & 0.5245 & 0.5360 & 0.4019 & 0.5026 & 0.5192 & 0.4405 & 0.3748 & 0.3661  & 0.3178 \\
        
        \bottomrule
    \end{tabular}
    \caption{\textbf{Ablation Studies of Different Components on T2I-CompBench and on our T2I-3DisBench.} Where T2I-CompBench uses the 3D-spatial metric because it is most relevant to the 3D environment, and T2I-3DisBench uses the average InternVL-VQA score in terms of object count, object orientation, 3D spatial relationship, and camera view.}
    \label{table: ablation}
    \vspace{-6px}
\end{table*}

\section{Experiment}

\noindent\textbf{Datasets and Metrics.}
We first conducted experiments on T2I-CompBench \citep{t2i-bench}
due to its comprehensive evaluations of object count and spatial relationships.
Since T2I-CompBench lacks detailed text prompts for object orientations and camera views. 
we further introduce T2I-3DisBench (3D image scene), 
a dataset of 50 textual prompts
that encapsulate complex 3D information.
We conducted both automatic and user evaluations on our T2I-3DisBench.
Since the metrics of T2I-compBench are inadequate for assessing detailed 3D features,
we uniquely employed Visual Question Answering (VQA) on InternVL \citep{internvl}.
Specifically,
we fed instructions to InternVL, asking it to score the generated images considering four dimensions: 
count, orientation, 3D spatial relationship, and camera view.
Additionally, 
we conducted user evaluation on our T2I-3DisBench,
where participants scored the images based on the same four dimensions.
More details about our T2I-3DisBench are shown in Appendix.

\noindent\textbf{Implementation Details.}
Our MUSES is modular and extensible,
allowing for integration of various LLMs, CLIPs, and ControlNets. 
In our experimental setup, 
we employed Llama-3-8B \citep{llama3} for 3D layout planning, 
ViT-L/14 for image/text encoding, 
ViT-B/32 for orientation calibration, 
and SD 3 ControlNet \citep{controlnet} for controllable image generation. 
For Llama-3-8B, 
we set top\_p to 0.1 and temperature to 0.2 
to ensure precise, consistent, and reliable outputs. 
For SD 3 ControlNet, 
we set inference steps to 20 and control scales from 0.5 to 0.9, 
as discussed in parameter ablation in Appendix. 
During evaluation, we select the best one from the five images of different control scales using benchmark metrics (e.g., T2I-CompBench and T2I-3DisBench).
The Blender's parameter conversion rules and settings are also specified in our Appendix. 
We use Mini-InternVL 1.5 \citep{far} for automated evaluation on T2I-3DisBench. 
Experiments are conducted on 8 NVIDIA RTX 3090 GPUs.

\noindent\textbf{SOTA Comparison on T2I-Compbench.}
As shown in Tab. \ref{table: ti2-combench-evaluation}, 
we compared our MUSES with two types of existing SOTA methods:
specialized/multi-agent approaches in the upper part
and 
generic models in the lower part. 
MUSES consistently outperforms both categories across all metrics, including object count, relationship, and attribute binding.
Specifically,
Our innovative 3D layout planning 
and 
3D-to-2D image conditions 
enhance object relationship understanding, 
leading to 
best performance on spatial-related metrics. 
Additionally, 
precise object positioning  boosts the numeracy score, %in Blender
and
detailed 3D model shape guidance significantly improves attribute binding scores.

\noindent\textbf{SOTA Comparison on T2I-3DisBench.}
As shown in Tab. \ref{table: complex-prompts-evaluation}, 
we conducted both automatic and user evaluations (in parentheses) on our T2I-3DisBench
and compared two types of SOTA methods as well.
\textit{For automatic evaluation on InternVL-VQA metrics},
our MUSES consistently outperforms others,
including the open-source SOTA model, 
Stable Diffusion V3, 
and closed-source API products 
like Midjourney v6.0.
Obviously, 
existing approaches struggle with 
complex prompts containing 3D information 
(e.g., object orientation, camera view),
highlighting the importance of our 3D-integration design.
\textit{For user evaluation},
We randomly selected 20 prompts from our T2I-3DisBench
and engaged 30 participants to rate image accuracy 
on a scale of 0.0 (poor) to 1.0 (excellent) 
across four dimensions.
The results show a strong preference for MUSES, 
demonstrating its effectiveness in handling complex 3D scenes.

\noindent\textbf{Ablation Studies.}
As shown in Table \ref{table: ablation}, 
our full system achieves the best performance. 
Removing any of the components will result in performance degradation.
\textit{(1) Object Depth Planning} is essential, as its removal leads to poor 3D spatial representation in Blender, affecting both datasets.
\textit{(2) Object Orientation Planning} is critical for prompts containing orientation information, with a score dropping on T2I-3DisBench when removed.
\textit{(3) Camera View Planning} affects performance mostly on T2I-3DisBench, since it contains camera information. 
\textit{(4) Retrieve-Generate Decision Tree} significantly influences performance on both datasets, 
highlighting the importance of high-quality 3D models. 
\textit{(5) Face-Camera Calibration} is the most important; its removal causes sharp performance drops as 3D models lose correct orientation specified in the 3D layout.
\textit{(6) CLIP Fine-tuning During Calibration} improves CLIP’s accuracy in determining object orientations, thus enhancing the orientation accuracy of our final image.
\textit{(7) Multiple Control Scales} effectively improves performance on both benchmarks.
\textit{(8) 3D Layout Shop Expansion} helps the LLM generate 3D and complex layouts better; it obviously works.
\textit{(9) Co-ablation of Multiple Components} (last three columns) 
shows that removing the Model Engineer has the most significant impact, resulting in poor object shaping and orientation.
Removing the Layout Manager also notably degrades performance.
Removing all components will result in the lowest score.
These findings demonstrate that each component is crucial for our 3D controllable image generation system.

\section{Conclusion}
We introduce MUSES, 
a multi-agent collaborative system for precise 3D controllable image generation. 
By integrating 3D layouts, models, and simulations, 
MUSES achieves fine-grained control over 3D object properties (e.g. object orientation, 3D spatial relationship) and camera view. 
To evaluate such complex 3D image scenes more comprehensively, we construct a new benchmark named T2I-3DisBench.
Experiments on T2I-CompBench and our new T2I-3DisBench demonstrate the superior performance of MUSES in handling complex 3D scenes. 
Future work will focus on 
improving efficiency and expanding capabilities to control lighting conditions, 
and potentially expanding to video generation.

\section{Acknowledgments}
This work was supported by the National Key R\&D Program of China(NO.2022ZD0160505), the National Natural Science Foundation of China under Grant(62272450), and the Joint Lab of CAS-HK.

\bigskip

\bibliography{aaai25}

\appendix

\section{NSR-1K Dataset Expansion}
\begin{figure}[H]
\centering
\includegraphics[width=1.0\columnwidth]{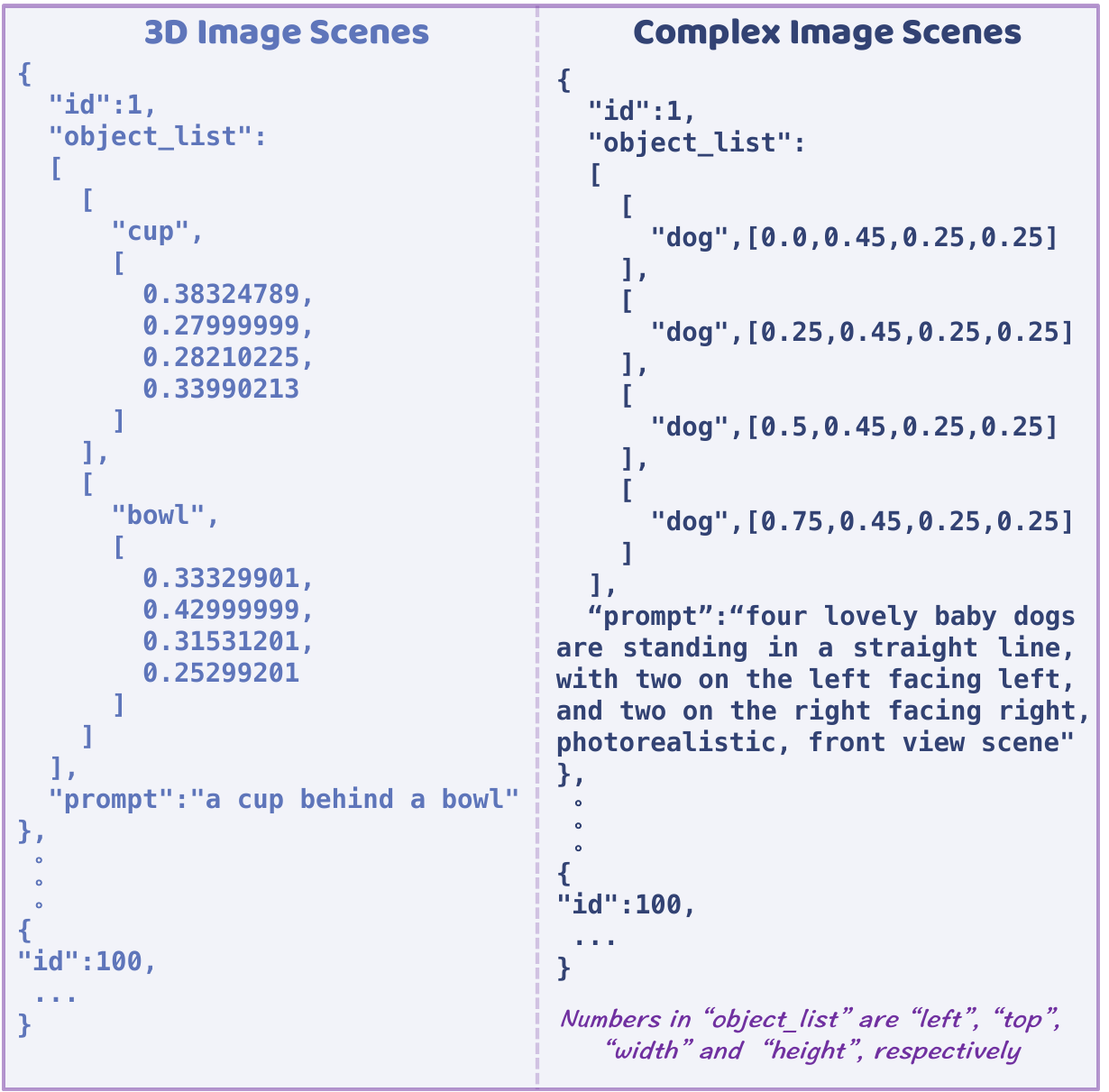}
\caption{\textbf{3D Layout Shop Expansion.} We extend the layout shop in two main aspects: 3D scene images and complex scene images. Each additional dataset entry contains attributes such as ``id", ``prompt", and ``object\_list".}
\label{fig6: expand_layout}
\end{figure}

Since the original layout dataset, NSR-1K \citep{t2i-bench} lacks detailed layouts in complex image scenes and 3D image scenes, we expand the dataset on these two types of image scenes. For each type of image scene, we manually construct 50 scenes and empirically design 3D layouts for these scenes. Then, we input the 50 pre-built text-layout pairs to Claude AI\footnote{https://www.anthropic.com/claude} in a context-learning manner and ask the LLM to generate another 50 for us. In this way, we obtain 100 text-layout pairs in complex scenes and 100 text-layout pairs in 3D scenes, respectively, and we add these 200 text-layout pairs to the original dataset to expand it. The NSR-1K-Expanded dataset has 40720 entries in total. 
Fig. \ref{fig6: expand_layout} shows the form of the text-layout pair we added in detail.

\begin{figure*}[ht]
\centering
\includegraphics[width=2.1\columnwidth]{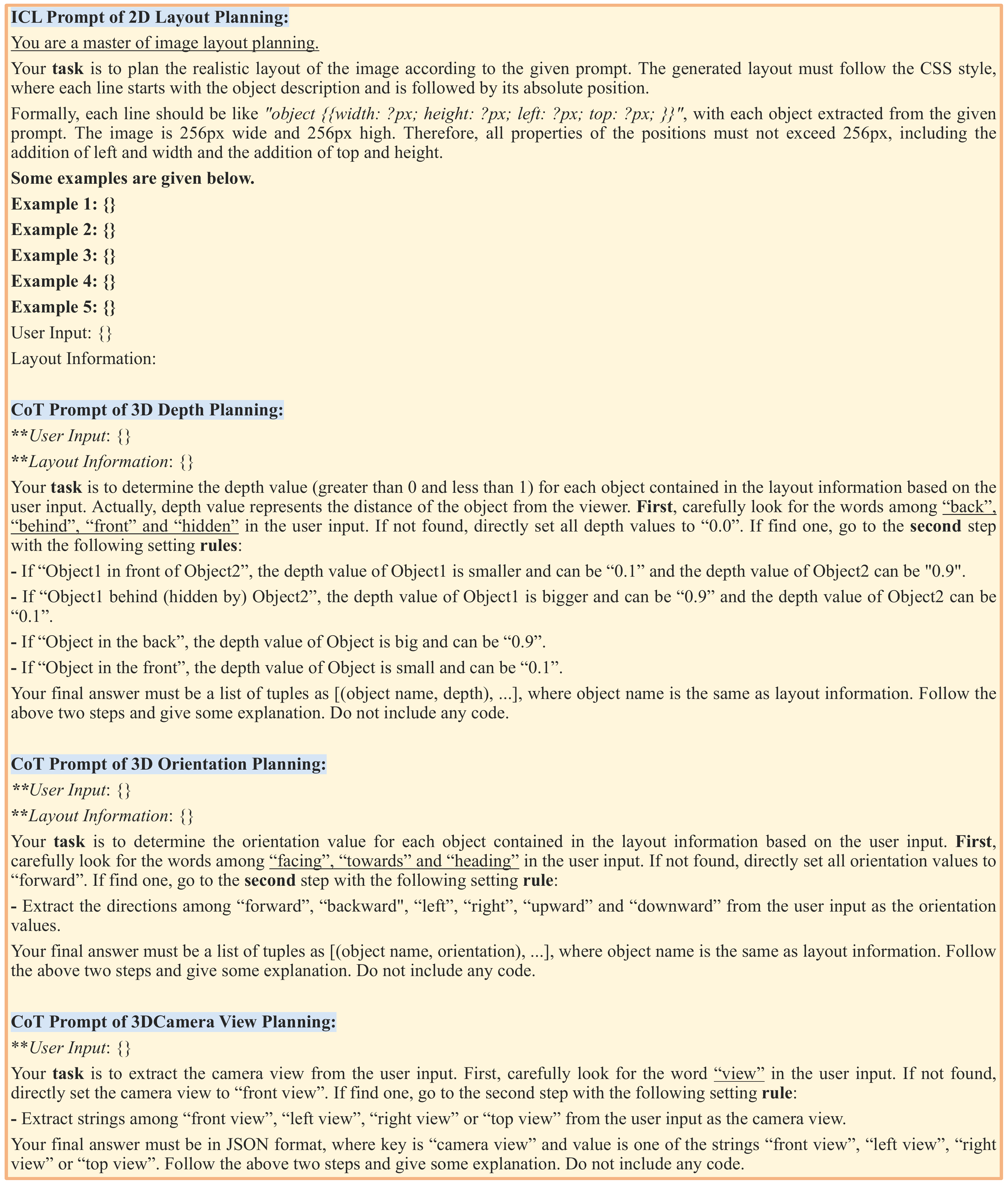}
\caption{\textbf{Full Planning Prompts in our 2D-to-3D Layout Manager.} There are four planning prompts in total, including ICL 2D layout planning, CoT 3D depth planning, CoT orientation planning, and CoT camera view planning.}
\label{fig7: prompt}
\end{figure*}

\section{Full Prompts for 3D Layout Planning}
\label{appendix:A}
We present our complete LLM planning prompts in Fig. \ref{fig7: prompt}, including 2D layout planning, 3D depth planning, orientation planning, and camera view planning. We fully take advantage of ICL and CoT prompting techniques to enhance LLM's reasoning and decision-making capabilities.

\begin{figure}[H]
\centering
\includegraphics[width=1.0\columnwidth]{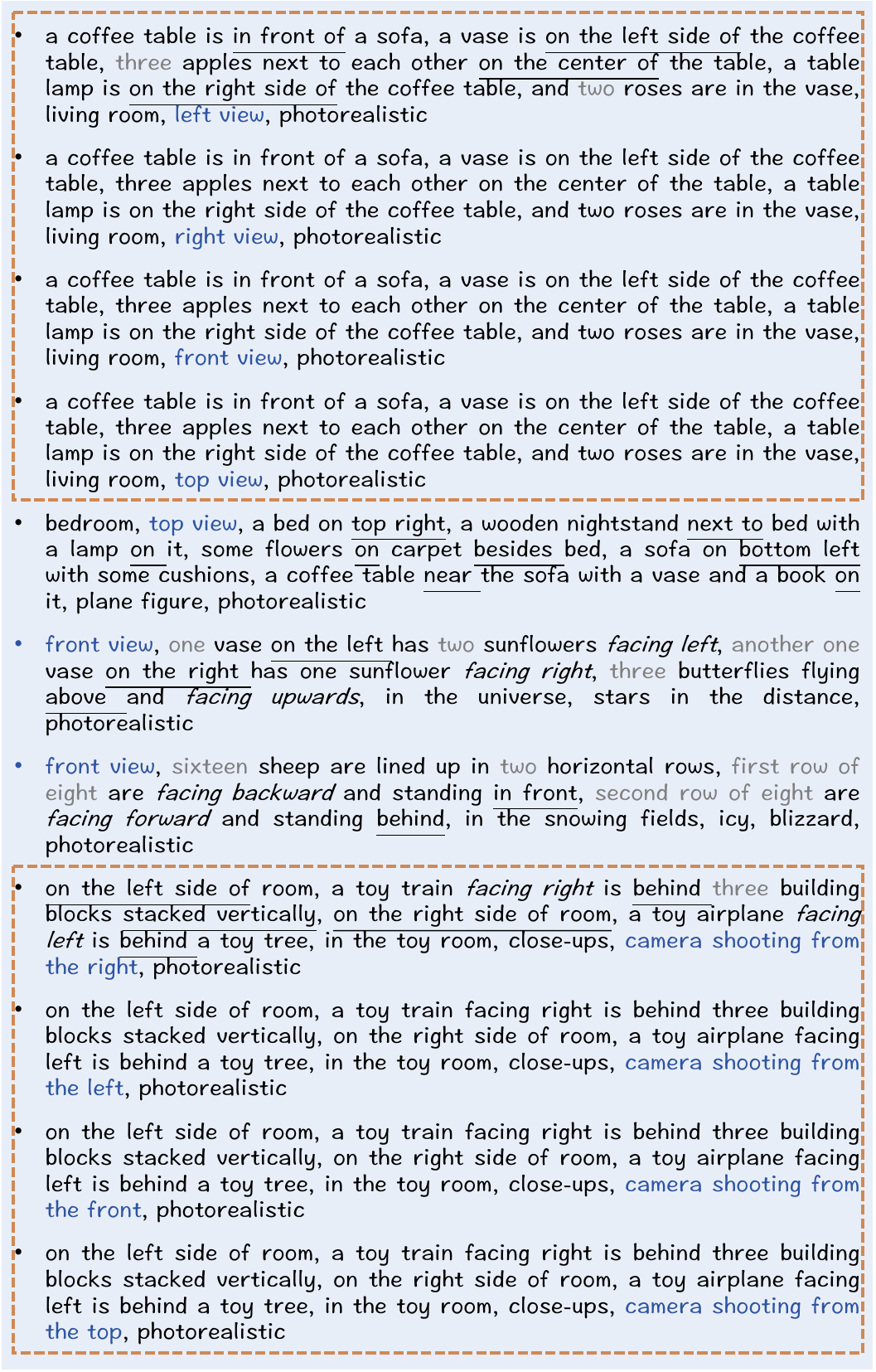}
\caption{\textbf{Representative Examples of Prompts in T2I-3DisBench.} Underlines indicate 3D spatial relationships, italics indicate object orientation, blue font indicates the camera view and gray font indicates the number of objects.}
\label{fig8: dataset}
\end{figure}

\section{T2I-3DisBench Benchmark}
\label{supp:B}

Owing to the absence of a suitable textual dataset involving multiple objects with various orientations, 3D spatial relationships, and camera views, we construct our benchmark, namely T2I-3DisBench (3D image scene Benchmark). The benchmark construction process begins with the careful crafting of 10 such complex and detailed prompts. To expand the dataset, we leverage the Claude AI as well to imitate and generate the remaining 40 similar texts, which are subsequently refined by human experts to ensure quality, diversity and consistency. Fig. \ref{fig8: dataset} presents some representative examples in our T2I-3DisBench textual dataset. For evaluation metrics, we find that traditional metrics such as CLIPScore \citep{clip} or BLIP-CLIP \citep{blip} lack the necessary precision to capture nuanced details like object orientations or camera views, and the metrics of T2I-compBench \citep{t2i-bench} are inadequate and inaccurate for assessing detailed 3D features.
Hence,
we employ InternVL \citep{internvl}, a Vision Large Language Model (VLLM), to score the generated images across four dimensions: 
object count, object orientation, 3D spatial relationship, and camera view. 
Specifically,
we feed the InternVL with the following instructions:

\emph{Text: \underline{\quad}.}

\emph{How well does the image match the text? You need to consider (1) object count, (2) object orientation, (3) 3D spatial relationship between objects, and (4) camera view.  Return a tuple ("score", X.XXXX), with the float number between 0 and 1, and higher scores representing higher text-image alignment.}

With such versatile instruction, we can comprehensively evaluate how well the generated images align with the text in terms of complex 3D details.

\section{Implementation Details in Blender}
\label{supplimen:C}
In this section, we provide a comprehensive overview of the procedures and code used to assemble 3D objects into a complete 3D scene and render it into a 2D image using Blender \citep{blender}. 
First, we need to initialize a Blender \texttt{bpy} environment, configuring global scene settings and rendering settings.
Then, we need to configure the camera parameters according to the planned 3D layout. Subsequently, we import all the 3D objects into the environment and set their corresponding parameters based on the planned 3D layout. Finally, we render the 3D scene into a 2D image and transform it into a depth map and edge map, which are later leveraged for fine-grained control in the ControlNet \citep{controlnet}. Our codes are all implemented on version 4.0.0 of Python Blender \texttt{bpy}.

\subsection{Environment Initialization}
To render a 3D scene into an image, we begin by initializing the Blender environment and setting up both the global scene and rendering parameters. 
For global scene settings, we set the background to a low-intensity gray (RGB: 0.1, 0.1, 0.1) using a shader node, creating a consistent gray backdrop. A global light source is positioned at (0, -5, 10) directly above the front of the object for uniform lighting across experiments.
For rendering settings, the Blender Cycles rendering engine is used for high-fidelity image output. Depth pass ($use\_pass\_z$) is enabled to extract depth information during rendering, and the color depth is set to 16-bit for better quality. The output resolution is configured to 1024x1024 pixels. The rendered image is saved as a PNG file in the specified directory. Our complete codes are shown in the following.

\vspace{10pt}
\begin{lstlisting}[language=Python, basicstyle=\scriptsize\ttfamily, breaklines=true, breakatwhitespace=true]
# Global Scene Settings
world = bpy.data.worlds['World']
world.use_nodes = True
bg_node = world.node_tree.nodes.get('Background')
if bg_node is None:
    bg_node = world.node_tree.nodes.new('ShaderNodeBackground')
bg_node.inputs[0].default_value = (0.1, 0.1, 0.1, 1)
bg_node.inputs[1].default_value = 1.0
light = bpy.data.objects['Light']
light.location = (0, -5, 10)
light.rotation_euler = (0, 0, 0)
# Rendering Settings
scene.render.engine = 'CYCLES'
scene.view_layers[0].use_pass_z = True
scene.render.image_settings.color_depth = '16'
scene.render.resolution_x = 1024
scene.render.resolution_y = 1024
scene.use_nodes = True
scene.render.filepath = os.path.join(output_img_dir, "rendering.png")
scene.render.image_settings.file_format = 'PNG'
\end{lstlisting}
\vspace{10pt}

\subsection{Camera Configuration}
After initializing the environment, we configure the camera parameters according to the 3D layout planned by our layout manager.
The parameters of the camera include the camera position and orientation, corresponding to $location$ and $rotation\_euler$ in \texttt{bpy}, respectively. We need to translate the camera view in the 3D layout into a parameterized form that Blender understands. For example, the camera position parameter (x,y,z) for “top view” is set to (0,1,15). Our full conversion rules are implemented as follows:
\vspace{10pt}
\begin{lstlisting}[language=Python, basicstyle=\scriptsize\ttfamily, breaklines=true, breakatwhitespace=true]
camera = bpy.data.objects['Camera']
if camera_view = "left view":
    randon_x = random.randint(-10, -5)
elif camera_view = "right view":
    randon_x = random.randint(5, 10)
else:
    randon_x = random.randint(-1, 1)
if camera_view = "top view":
    camera.location = (0, 1, 15)
else:
    camera.location = (randon_x, -math.sqrt(225 -       randon_x**2), 5)
camera.rotation_euler = (math.atan(math.sqrt(camera_x**2 + camera_y**2) / (camera_z + 1e-5)), 0, -math.atan(camera_x / (camera_y + 1e-5)))
\end{lstlisting}
\vspace{10pt}

\subsection{Object Parameter Settings}
Next, we import all the 3D objects (OBJ files) into our environment using the $bpy.ops.wm.obj\_import(filepath='')$ function. For each 3D model, we set key parameters, including object size, position, and orientation, which correspond to the $dimensions$, $delta\_location$, and $rotation\_euler$ attributes in \texttt{bpy}, respectively. The complete code for our comprehensive conversion process is provided below: 

\vspace{10pt}
\begin{lstlisting}[language=Python, basicstyle=\scriptsize\ttfamily, breaklines=true, breakatwhitespace=true]
# Load JSON of 3D Layout
with open('3D_layout_info.json', 'r') as file:
    layout = json.load(file)
# Load JSON of Alignment Imformation
with open('3D_Model_Aligner_info.json', 'r') as file:
    align = json.load(file)
# Import i-th OBJ File
bpy.ops.wm.obj_import(filepath=obj_files[i])
# Merge Extra Components
if (len(bpy.data.objects) > 3 + i):
    obs = bpy.context.scene.objects[2:]
    ctx = bpy.context.copy()
    ctx['selected_objects'] = obs
    bpy.ops.object.join()
obj = bpy.context.selected_objects[0] 
# Set 'dimensions'
height_width = obj.dimensions[2] / obj.dimensions[0]
height_depth = obj.dimensions[2] / obj.dimensions[1]
width_depth = obj.dimensions[0] / obj.dimensions[1]
max_obj_size = max(obj.dimensions[0], obj.dimensions[2])
max_item_size = max(layout[i]['width'], layout[i]['height'])*10 + layout[i]['depth']
if max_obj_size == obj.dimensions[0]:
    obj.dimensions = (max_item_size, max_item_size / width_depth, max_item_size * height_width)
else:
    obj.dimensions = (max_item_size / height_width, max_item_size / height_depth, max_item_size)
bpy.ops.object.transform_apply(location=False, rotation=False, scale=True)
bpy.context.view_layer.update()
if item_data[i]['depth'] < 0.5:
    while sum(obj.dimensions) > 25:
        obj.dimensions = obj.dimensions / 1.5
        bpy.context.view_layer.update()
# Set 'delta_location'
bpy.ops.object.origin_set(type='ORIGIN_CENTER_OF_MASS', center='BOUNDS')
obj.location = (0, 0, 0)
obj.delta_location[0] = (layout[i]['left'] + layout[i]['width']/2 - 0.5) * 10* (0.9 + layout[i]['depth'])
obj.delta_location[1] = (layout[i]['depth']-0.1)*(10+obj.dimensions[1])
obj.delta_location[2] = -(layout[i]['top'] + layout[i]['height']/2 - 0.5) * 10 * (0.9 + layout[i]['depth']) - 3 * layout[i]['depth']
bpy.ops.object.transform_apply(location=True, rotation=False, scale=False)
bpy.context.view_layer.update()
# Set 'rotation_euler'
rotation = {"forward": [0, 0, 0], "backward": [0, 0, math.pi], "left": [0, 0, -90/180 * math.pi], "right": [0, 0, 90/180 * math.pi], "upward": [-90/180 * math.pi, 0, 0], "downward" : [90/ 180 * math.pi, 0, 0]}
obj.rotation_euler = [x+y for x, y in zip(align[i], rotation[layout[i]['orientation']])]
bpy.ops.object.transform_apply(location=False, rotation=True, scale=False)
bpy.context.view_layer.update()
\end{lstlisting}
\vspace{10pt}

These conversion rules ensure accurate translation of the planned 3D layout into Blender's 3D simulation environment, referencing the face-camera-view orientation of the object as determined by the 3D model engineer. Through such comprehensive parameter configurations, we achieve high alignment of object placement with the planned 3D layout. This is significant to the accuracy of the object orientation in the final generated image.

\subsection{Image Rendering Settings}
After setting all the parameters, we now successfully assemble the 3D scene. We can render the complete 3D scene into a 2D image with accuracy using the $bpy.ops.render.render(True)$ function. This 2D rendering is further transformed into both a depth map and an edge map, which are used for fine-grained control in the ControlNet \citep{controlnet}.
To generate the depth map, we use a depth node within Blender to capture the Z-depth values of the 3D objects. This depth information is crucial for understanding the spatial relationships between objects in the scene.
For the edge map, we employ OpenCV to detect the contours in the rendered image. This edge map highlights the boundaries and shapes of objects, providing additional information that aids in the precise manipulation of the 3D scene. Codes are presented in the following.

\vspace{10pt}
\begin{lstlisting}[language=Python, basicstyle=\scriptsize\ttfamily, breaklines=true, breakatwhitespace=true] 
# Depth Map
tree = scene.node_tree
links = tree.links
# Clear Nodes
for n in tree.nodes:
    tree.nodes.remove(n)
render_layers_node = tree.nodes.new(type='CompositorNodeRLayers')
invert_node = tree.nodes.new(type='CompositorNodeInvert')
depth_output_node = tree.nodes.new(type='CompositorNodeOutputFile')
depth_output_node.base_path = output_img_dir + "/depth"
depth_output_node.file_slots[0].path = "depth000"
map_node = tree.nodes.new('CompositorNodeMapValue')
map_node.offset = [-20]
map_node.size = [0.1]
links.new(render_layers_node.outputs[2], map_node.inputs[0])
links.new(map_node.outputs[0], invert_node.inputs[1])
links.new(invert_node.outputs[0], depth_output_node.inputs[0])
# Rendering Image
bpy.ops.render.render(write_still=True)
# Edge Map
import cv2
image = cv2.imread(os.path.join(output_img_dir, "rendering.png"))
edges = cv2.Canny(image, 100, 200)
cv2.imwrite(os.path.join(output_img_dir, "canny_edges.png"), edges)
\end{lstlisting}
\vspace{10pt}

The meticulous setup of the environment and the precise configuration of camera parameters and object parameters collectively contribute to high-quality rendering. This high-quality rendering is essential for accurately and effectively controlling the 3D properties of the generated image. As a result, it greatly facilitates our precise and reliable 3D-controllable image generation in the MUSES system.

\vspace{5pt}
\section{Ablation of ControlNet Parameters}
\label{appendix:D}
\begin{figure*}[ht]
\centering
\includegraphics[width=2.13\columnwidth]{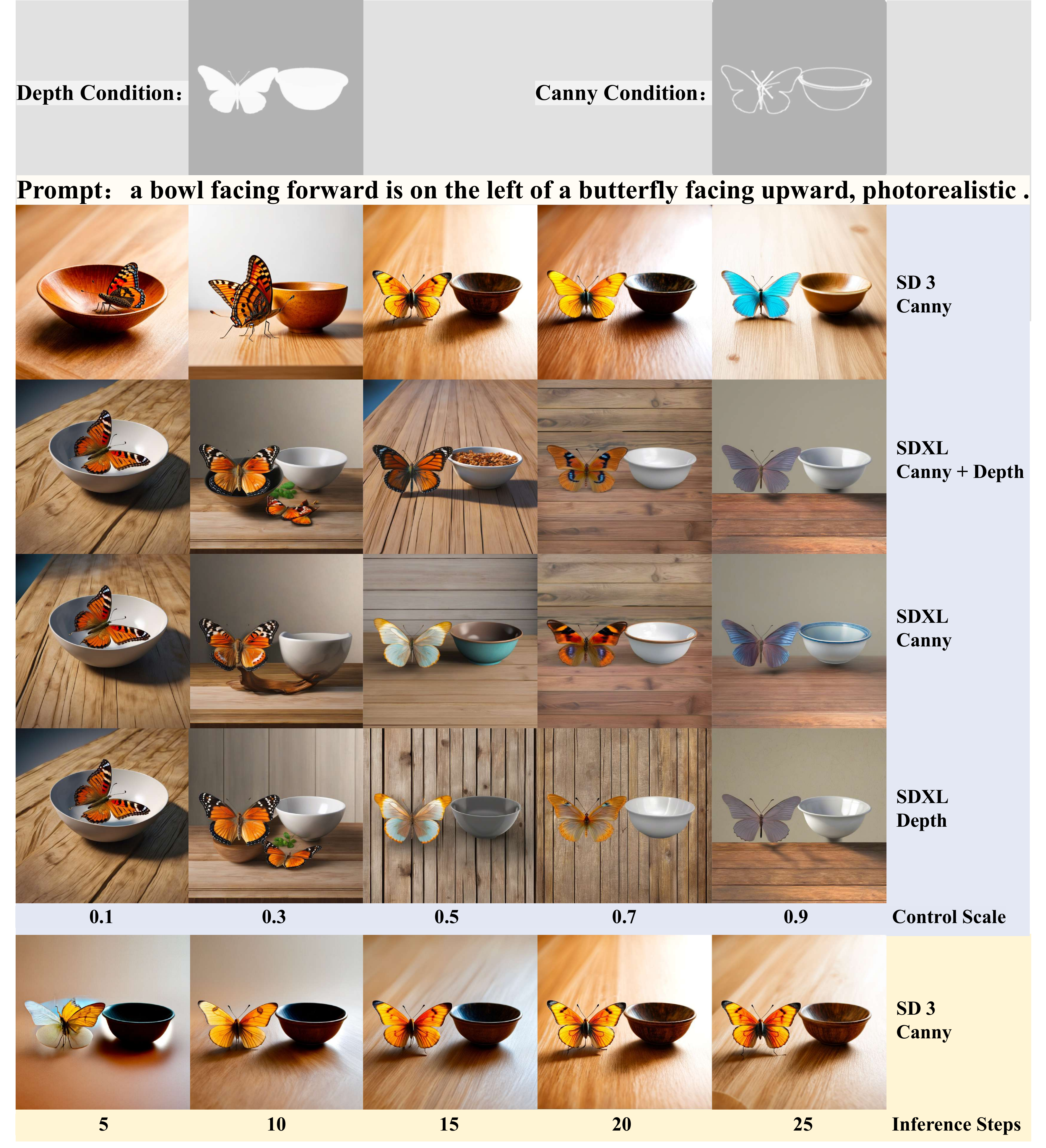}
\caption{\textbf{Qualitative Comparison with Different Inference Steps and Control Scale Parameters of ControlNets.} The larger the control scale, the higher the alignment of the generated image with the input conditions, but the lower the image fidelity. Meanwhile, as the inference steps increase, the image quality improves until it is saturated (after 20 inference steps).}
\label{fig9: ablation_param}
\end{figure*}

We conduct comparative experiments with different control scales and inference steps of ControlNets to determine the optimal parameter settings. The $control\ scale$ parameter ranges from 0.1 to 1.0, and the $inference\ steps$ parameter ranges from 5 to 30. As shown in Fig. \ref{fig9: ablation_param}, we select several representative parameter values and visualize the comparison results.
The results indicate that increasing the control scale enhances the alignment of the generated images with the input condition images. 
When the control scale reaches 0.5, the details of the generated image can already be well controlled. It is difficult to determine which control scale results in images with better quality after 0.5. 
Therefore, during the experiments, we set the control scale of SD 3 ControlNet in the interval of [0.5, 0.9]. That is, each run generates five images with different control scales and then evaluates them using benchmark metrics (e.g., T2I-CompBench and T2I-3DisBench) to select the best image with the highest score.

In terms of inference steps, the quality of the generated images improves with an increasing number of steps. However, once the number of steps exceeds 20, the image quality plateaus. Therefore, 20 inference steps are chosen as the optimal inference steps in our experimental setup.

\vspace{5pt}
\section{More Qualitative Comparisons}
\label{appendix:E}
Figure \ref{fig10: complex_prompts_demo1} and Figure \ref{fig11: complex_prompts_demo2} present more qualitative comparisons between our MUSES and existing state-of-the-art methods, including both open-source models and commercial API products, such as Stable Diffusion V3 \citep{sd3}, DALL-E 3 \citep{dalle3}, and Midjourney v6.0 \citep{midjourney}. Our systematic collaborative approach generates images that are more faithful to the details of the text, in terms of object count, orientation, 3D spatial relationships, and camera view. Even state-of-the-art methods fail to generate precise images that accurately represent the complex 3D details in the input text prompts.

\vspace{5pt}
\section{More Blender Rendering Outputs}
To further demonstrate the rendering quality of our approach, 
we present additional rendering outputs generated using Blender in Fig. \ref{fig15: blender_rendering}.
Specifically,
we import 3D models into Blender according to the planned 3D layout and then render the 3D scene into a 2D image via bpy's “CYCLES” Rendering Engine.
These renderings can be further transformed into depth maps and canny edges, illustrating the controllability of our novel MUSES pipeline.

\vspace{5pt}
\section{Limitation}
\begin{figure}[H]
\centering
\includegraphics[width=1.0\columnwidth]{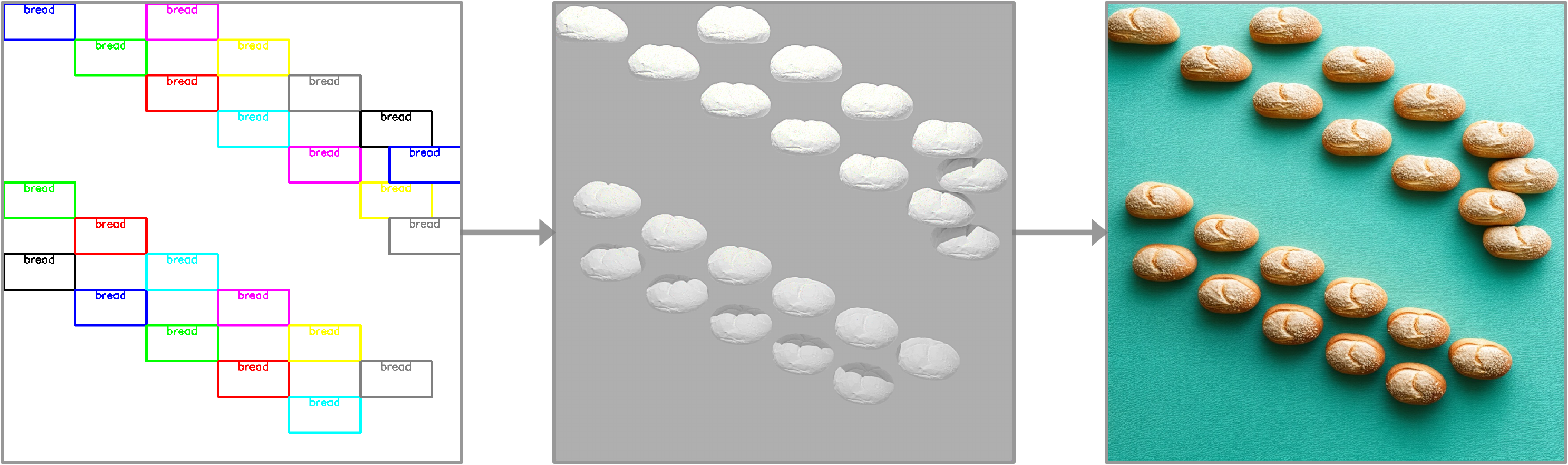}
\caption{\textbf{Failure Case.} The user input prompt is ``twenty five bread". Since this prompt involves a large number of objects without explicit information on object placement, it is difficult for Layout Manager to generate a Visually good 3D layout. On the contrary, the result is better if the input prompt is more detailed (``Twenty five bread arranged in five horizontal lines, fill the entire picture, on the white carpet, top view.") in Fig. \ref{fig1: simple_prompts_demos}.}
\label{fig13: failure_case}
\end{figure}

\begin{figure*}[t]
\centering
\includegraphics[width=2.15\columnwidth]{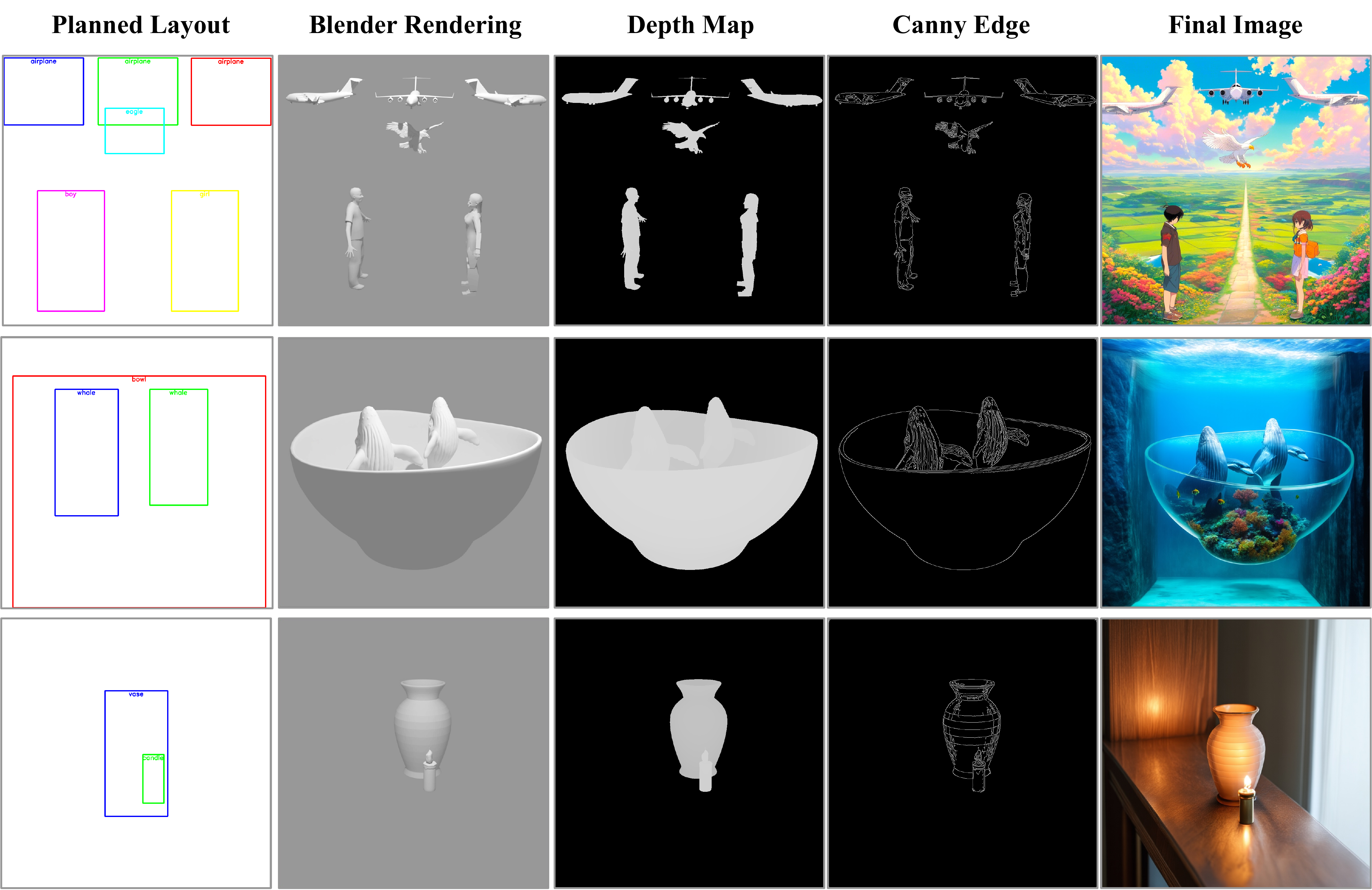}
\caption{\textbf{More Rendering Outputs.} The user prompt of ROW 1 is ``\textit{three airplanes flying in the air, facing left, forward, right, respectively, an eagle flying underneath the center airplane, a boy facing right stands left, a girl facing left stands right, on a beautiful path stretching into the distance, colorful scene, anime style}". The user prompt of ROW 2 is ``\textit{two blue whales, swimming upwards in a huge fish bowl, surrounded by colorful coral and small fish swimming around them, left view, sci-fi, beautiful scene, complex scene, photorealistic}". The user prompt of ROW 3 is ``\textit{a vase facing forward is behind a candle facing backward, placed on a polished wooden mantelpiece, the candle, which is tall and elegant, is lit and casts a warm, flickering glow, front view, quiet atmosphere, photorealistic}".}
\label{fig15: blender_rendering}
\end{figure*}

\begin{figure*}[t]
\centering
\includegraphics[width=2.15\columnwidth]{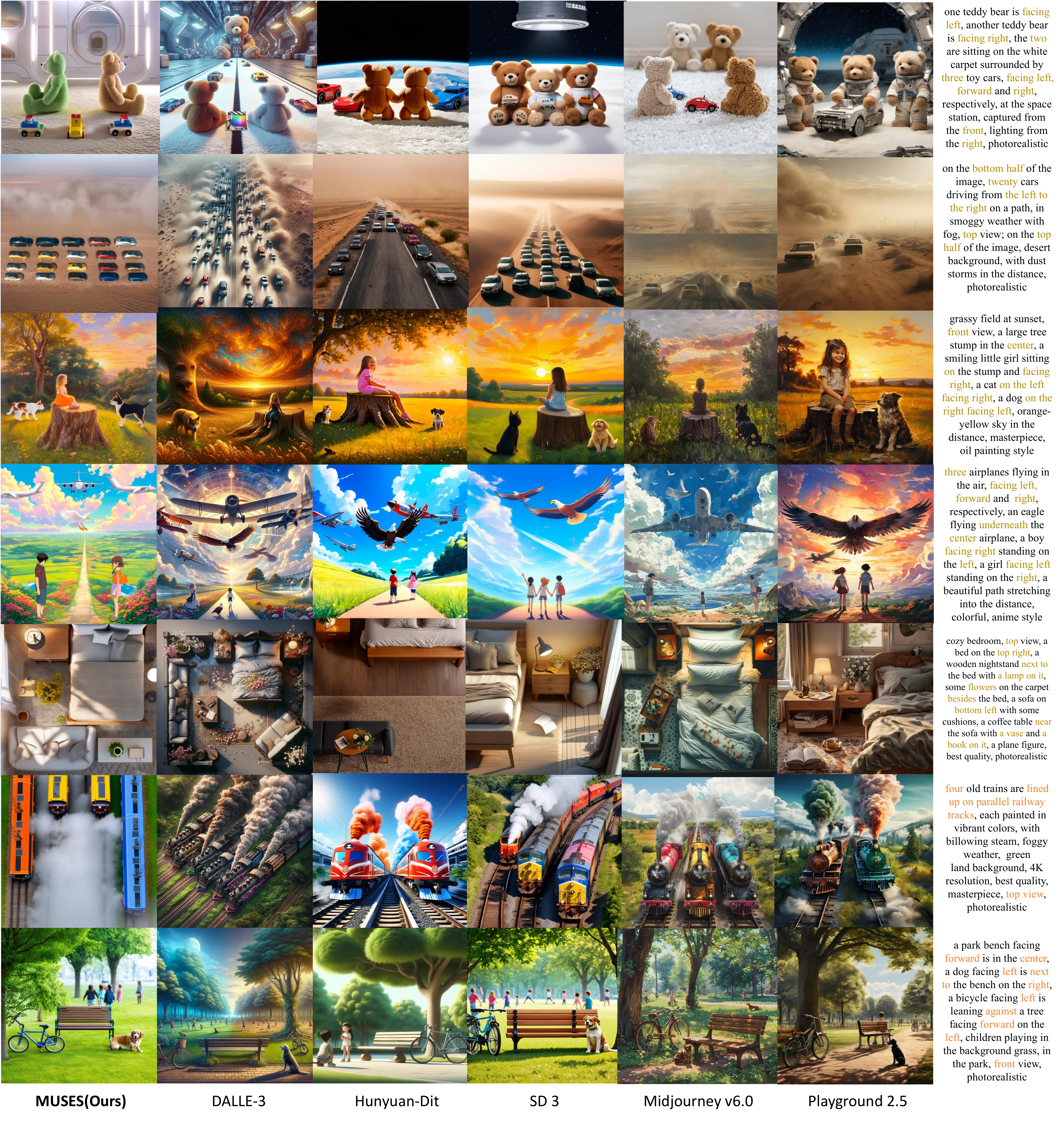}
\caption{\textbf{Qualitative Comparison (1) With Existing SOTA Methods.} Our MUSES consistently demonstrates the highest text-image alignment, regarding object count, object orientation, 3D spatial relationship between objects, and camera view. We compare our MUSES with SOTA open-sourced models and closed-sourced commercial API products, including Stable Diffusion V3, Hunyuan-Dit, Playground v2.5, DALL-E 3, and Midjourney v6.0. \textcolor{orange}{Color-marked texts} on the far right represent important 3D details.}
\label{fig10: complex_prompts_demo1}
\end{figure*}

\begin{figure*}[t]
\centering
\includegraphics[width=2.15\columnwidth]{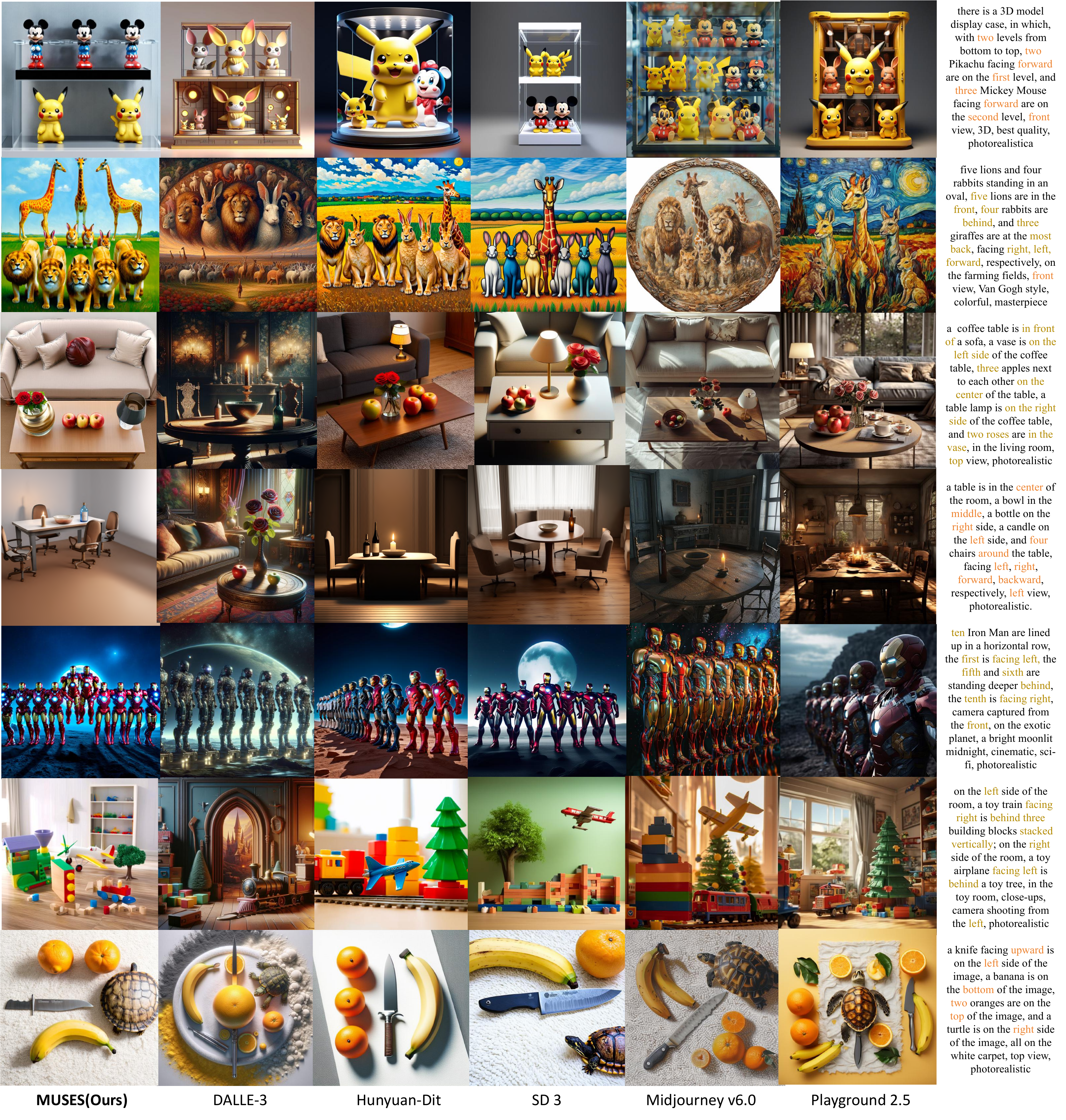}
\caption{\textbf{Qualitative Comparison (2) With Existing SOTA Methods.} Our MUSES consistently demonstrates the highest text-image alignment, regarding object count, object orientation, 3D spatial relationship between objects, and camera view. We compare our MUSES with SOTA open-sourced models and closed-sourced commercial API products, including Stable Diffusion V3, Hunyuan-Dit, Playground v2.5, DALL-E 3, and Midjourney v6.0. \textcolor{orange}{Color-marked texts} on the far right represent important 3D details.}
\label{fig11: complex_prompts_demo2}
\end{figure*}

Although our MUSES achieves precise 3D properties control of image generation (e.g., 3D layout, orientation, camera view). Our efficiency is relatively low compared to methods like SD3, since our pipeline involves multi-agent collaboration. 
In addition,
we have observed that MUSES would fail for user prompts that contain a large number of objects with unclear layout descriptions. 
Taking Fig. \ref{fig13: failure_case} as an example,  
these situations can introduce ambiguity that affects the precision of object placement and orientation in generated scenes.
Therefore, we highly recommend that users
either provide a detailed prompt or self-define the 3D layout of each object for better image generation results.

\end{document}